\documentclass{article}

\usepackage{PRIMEarxiv}

\usepackage[utf8]{inputenc} 
\usepackage[T1]{fontenc}    
\usepackage{hyperref}       
\usepackage{url}            
\usepackage{booktabs}       
\usepackage{amsfonts}       
\usepackage{nicefrac}       
\usepackage{microtype}      
\usepackage{lipsum}
\usepackage{fancyhdr}       
\usepackage{graphicx}       
\graphicspath{{media/}}     

\usepackage{algorithm}
\usepackage{algpseudocode}
\usepackage{wrapfig}
\usepackage{subcaption}
\usepackage{enumerate}
\usepackage{color,array}
\usepackage{amsmath}

\title{BEExformer: A Fast Inferencing Binarized Transformer with Early Exits
\thanks{\textit{\underline{Citation}}: 
\textbf{Ansar, Wazib, Saptarsi Goswami, and Amlan Chakrabarti, "BEExformer: A Fast Inferencing Binarized Transformer With Early Exits," in IEEE Transactions on Sustainable Computing, vol. 11, no. 2, pp. 98-110, 2026, https://doi.org/10.1109/TSUSC.2026.3666456.} \\
\copyright 2026 IEEE. Personal use of this material is permitted. Permission from IEEE must be obtained for all other uses, in any current or future media, including reprinting/republishing this material for advertising or promotional purposes, creating new collective works, for resale or redistribution to servers or lists, or reuse of any copyrighted component of this work in other works.} 
}

\author{
    Wazib~Ansar \\
  A. K. Choudhury School of IT \\
  University of Calcutta \\
  Kolkata, West Bengal, India \\
  \texttt{waakcs\_rs@caluniv.ac.in} \\
   \And
  Saptarsi~Goswami \\
  Department of Computer Science \\
  Bangabasi Morning College \\
  Kolkata, West Bengal, India \\
  \texttt{sgakc@caluniv.ac.in} \\
    \And
  Amlan~Chakrabarti \\
  A. K. Choudhury School of IT \\
  University of Calcutta \\
  Kolkata, West Bengal, India \\
  \texttt{acakcs@caluniv.ac.in} \\  
}

\begin{document}
\maketitle

\begin{abstract}
Large Language Models (LLMs) based on transformers achieve cutting-edge results on a variety of applications. However, their enormous size and processing requirements hinder deployment on constrained resources. To enhance efficiency, binarization and Early Exit (EE) have proved to be effective solutions. However, binarization may lead to performance loss as reduced precision affects gradient estimation and parameter updates. Besides, research on EE mechanisms is still in its early stages. To address these challenges, we introduce Binarized Early Exit Transformer (BEExformer), a first-of-its-kind selective learning-based transformer integrating Binarization-Aware Training (BAT) with EE for efficient and fast textual inference. Each transformer block has an integrated Selective-Learn Forget Network (SLFN) to enhance contextual retention while eliminating irrelevant information. The BAT employs a differentiable second-order approximation to the sign function, enabling gradient computation that captures both the sign and magnitude of the weights. This aids in 21.30 $\times$ reduction in model size. The EE mechanism hinges on fractional reduction in entropy among intermediate transformer blocks with soft-routing loss estimation. This accelerates inference by reducing FLOPs by 52.27\% and even improves accuracy by 3.22\% by resolving the ``overthinking" problem inherent in deep networks. Extensive evaluation through comparison with the SOTA methods and various ablations across nine datasets covering multiple NLP tasks demonstrates its Pareto-optimal performance-efficiency trade-off.

\end{abstract}


\section{Introduction}
The last decade has witnessed substantial progress in the field of Natural Language Processing (NLP). The current state-of-the-art (SOTA) comprises pre-trained Large Language Models (LLMs) involving transformers \cite{li2022survey}. However, such models have exorbitant computational complexity, leading to high energy consumption and carbon emissions \cite{Ref41, Ref42}. This necessitates the need for sustainable practices focusing on efficient and economical use of computing resources \cite{Ref43, wa2025survey}. Efficient modeling reduces the memory usage and the number of computations, leading to faster inference. Various modeling aspects have been considered to enhance the efficiency of models, including pruning \cite{Ref21}, knowledge distillation (KD) \cite{Ref22}, low-rank approximation \cite{Ref23}, early exit (EE) \cite{Ref6, Ref26, Ref27} and model quantization \cite{Ref12, Ref29, Ref19}.

Of these, quantization has become quintessential for the deployment of NLP models with constrained resources \cite{Ref24}. Unlike most efficiency considerations that aim to reduce the number of layers, nodes, or parameters; quantization lowers the precision of the model’s memory representation \cite{rokh2023comprehensive}. In neural networks, quantization can be segregated as Post Training Quantization (PTQ) \cite{Ref29, Ref30, Ref31} wherein weights and/or activations are quantized post-training and Quantization Aware Training (QAT) \cite{Ref19, Ref25} which integrates quantization during the training process itself. QAT performs quantization of the weights and/or activations during the forward pass. During backpropagation, it uses full-precision latent weights and computes pseudo gradients to deal with zero gradient problem due to quantization. The class of neural networks employing 1-bit quantization is referred to as Binarized Neural Networks (BNN) and QAT in context to BNN is called Binarization-Aware Training (BAT) \cite{qin2023bibench}. It is an extreme form of quantization with 32x lower precision compared to float32 \cite{Ref18}. Although BNN seems to be a promising efficiency enhancement approach, only a handful of research works have been able to effectively implement it in the field of NLP \cite{Ref1, Ref2, Ref3}. Most of them have attempted BNN through KD from a full-precision LLM like BERT \cite{Ref1, Ref2}. Even though KD leads to a lightweight model, the training process itself is computationally as well as memory intensive \cite{Ref28}. Besides, BNNs with only single-bit weights and activations, have severely limited representational capacity, making it hard to mimic a full-precision teacher \cite{Ref46}. To the best of our knowledge, none of the works have attempted to devise a binarized transformer architecture from scratch. 

Along with quantization, dynamic variations in the model architecture accelerate inference by optimizing computational requirements based on input \cite{Ref44}. However, limited efforts have been directed towards making inference faster \cite{Ref40}. In LLMs, an input sequence is processed through multiple transformer blocks to generate the final output. However, heterogeneity can be observed in the inputs due to varying sequence lengths and sentence complexities \cite{Ref6}. Some inputs might require processing through all layers. For others, the lower or intermediate layers might be sufficient to yield predictions beyond a certain level of confidence. In such cases, while predictions based on lower layers are correct, they might become incorrect after undergoing transformations in the higher layers due to computation of additional parameters. This leads to wasteful computation or ``overthinking.” ``Overthinking" is an inference-time, input-dependent phenomenon wherein the model continues processing through additional parameters beyond the point needed to produce a correct output leading to higher FLOPs and potential performance degradation \cite{Ref4, Ref6}. Therefore, allowing EE from multiple break-points based on certain conditions provides a remedy to ``overthinking" \cite{Ref5, Ref6, Ref26} and even enhances efficacy \cite{Ref27}. The previous research on EE mechanisms for NLP has been on models with weights initialized from a pre-trained LLM \cite{Ref5, Ref6, Ref14}. The EE criterion is often a set threshold that needs to be calibrated based on the task as well as the input distribution \cite{Ref5}. This leads to faltering efficacy when inputs deviate during inference. Despite the potential for improved efficiency, no prior work has combined BAT with EE in a transformer architecture for textual inference. A key challenge in integrating EE with BNN is that it tends to destabilize training, leading to vanishing gradients. Additionally, standard loss functions are unsuitable for such networks.

To mitigate these challenges, we propose BEExformer-- a first-of-its-kind binarized transformer architecture with selective learning and EE pathways for textual inference. The integration of BAT and EE offers a dual advantage of a lightweight model with reduced precision, and accelerated inference by dynamically skipping transformer blocks based on output confidence. It comprises an intuitive BAT mechanism with real-valued latent weights leveraging a differentiable second-order approximation to the sign function. This ensures that the gradient is aware of the sign and magnitude of the weights. For EE, fractional change in the entropy of the logits is estimated after each transformer block during inference. If the entropy does not decrease beyond a threshold percentage, the exit condition is satisfied, and the final logits are returned. This reduces FLOPs by bypassing the execution of blocks beyond the exit point. Additionally, it mitigates performance degradation due to ``overthinking". During training, soft-routing loss is applied, combining the loss from all exits to effectively update the weights and enhance the decision-making capability of each transformer block. Moreover, each transformer block has a binarized version of the Selective Learn-Forget Network (SLFN) \cite{Ref15} integrated in it, leading to the elimination of insignificant information and better comprehension of long-range dependencies. Furthermore, BEExformer learns its parameters from scratch. This reduces the additional computational and memory requirements associated with using a full-precision LLM as a starting point. All this contributes towards Pareto-optimal performance-to-complexity trade-off compared to the SOTA on datasets like SST-2, CoLA, MRPC, RTE, MNLI-m, MNLI-mm, STS-B, QNLI and QQP. The principal contributions of this paper are as follows:

\begin{enumerate}
    \item We propose BEExformer-- a first-of-its-kind transformer architecture trained from scratch, combining BAT with EE for textual inference. Additionally, each transformer block is integrated with binarized SLFN to enable selective learning by eliminating insignificant information.
    
    \item For BAT, we propose a mechanism for binarizing weights and activations of transformer architecture through a tight second-order approximation to the sign function. This makes it differentiable and ensures that gradient-based weight updates consider both magnitude and sign of the weights, leading to 21.30 × reduction in model size.
    
    \item For EE, we propose a mechanism based on fractional entropy changes in logits across transformer blocks, eliminating the need for absolute thresholds. This is accompanied by a soft-routing loss estimation during training to enhance the inferential ability of each transformer block. It reduces FLOPs by 52.27\% and even improves accuracy by 3.22\% through mitigating ``overthinking".
    
\end{enumerate}

The compendium to this paper has been presented herein. Section \ref{rw} gives an overview of the related works in the domain. Section \ref{pm} describes the proposed BEExformer architecture. Section \ref{es} contains the experiment details. Section \ref{rd} discusses the results obtained through extensive evaluation on multiple datasets. Finally, conclusions are drawn and the future scope is envisioned in Section \ref{cl}.

\section{Related Works}
\label{rw}

In this section, a commentary on the related works has been presented. Since BEExformer is the first model combining BNN with EE for textual inference, the works on BNN and EE have been discussed separately.

\subsection{Binarized Neural Networks (BNN)}
\label{brw}
Shen et al. \cite{Ref12} proposed a mixed-precision, group-wise quantization method for BERT, using second-order Hessian insights to optimize compression. By analyzing Hessian eigenvalues, they refined precision allocation, while group-wise quantization enhanced control within attention layers by treating each dense matrix as a distinct group. Zhang et al. \cite{Ref13} attempted ternerization of BERT through combined approximation and loss-aware approach. Bai et al. \cite{Ref1} implemented weight binarization in a BERT model. To tackle sharp loss in performance, they apply KD along with ternary-weight splitting to initialize their model. However, they experience hindrances in binarizing activations and were not able to implement a fully binarized network due to a sharp performance drop due to loss of precision. Qin et al. \cite{Ref2} proposed a fully binarized BERT model applying binarization subject to information entropy maximization along with direction-matching distillation inferring from similarity matrices to deal with direction mismatch. Liu et al. \cite{Ref19} formulated a tight approximation to the sign function and updated the weights based on a magnitude-aware gradient. Liu et al. \cite{Ref3} further proposed a multi-step distillation approach to gradually distill the model to lower precision before obtaining the final BNN from a full-precision BERT model. They utilize an elastic binary activation with parameters inferred while training. All of the above-mentioned works adopt KD from a full-precision model to counter performance loss. However, executing both student and teacher models simultaneously for KD can be memory-intensive. Furthermore, additional fine-tuning on downstream tasks is often required for optimal results.

\subsection{Early Exit (EE)}
\label{erw}
Fan et al. \cite{fan2019reducing} introduced a structured dropout technique for transformers. By selectively dropping layers, it allowed sub-networks of varying depths to be extracted from a single trained model without requiring fine-tuning. Michel et al. \cite{michel2019sixteen} applied structured pruning to selectively remove less impactful attention heads, leading to a more efficient model with a significant reduction in inference costs. Later, Zhou et al. \cite{Ref4} developed an EE mechanism with the layers of a Pre-trained Language Model (PLM) interlaced with internal classifiers to exit from the network. They determined the change in predictions of successive internal classifiers and the exit criterion was fulfilled if no change could be observed for a given number of steps. In the same context, Xin et al. \cite{Ref5} calculated the entropy of the predictions of the internal classifiers, and if it fell short of a predetermined threshold, the exit criterion was fulfilled. However, determining the optimal threshold value is a daunting task. Liu et al. \cite{Ref14} put forth a pre-trained transformer model with dynamic exits having loss calculated at each layer as the sum of Masked-Language Modeling (MLM) and Sentence-Order Prediction (SOP) losses. Mangrulkar et al. \cite{Ref6} proposed a modified version of the switch transformer, formulating a switch layer to assess the complexity of a sentence and optimally route it to a set of expert models varying in number of encoder layers. However, the effectiveness of routing hinges on the capability of experts to deal with varying input patterns.

\subsection{Challenges of Integrating BNN with EE}
\label{berw}
It is observed that none of the works to the best of our knowledge have integrated BNN with EE in a transformer architecture for textual inference. This can be attributed to the multiple challenges involved. Binarization reduces weight and activation precision, destabilizing gradient propagation. Introducing EE complicates this further due to multiple exit points that require careful optimization to prevent vanishing gradients. Moreover, standard loss functions cannot be applied to a network with both BNN and EE, necessitating a tailored loss that preserves accuracy by enhancing the inferential capability of each exit point. Finally, estimating the EE confidence threshold in a BNN is a challenging task.

\section{Proposed Architecture}
\label{pm}

\begin{figure*}[h!]
    \centering
    \begin{subfigure}[h]{\linewidth}
        \centering
        \includegraphics[width=0.96\linewidth]{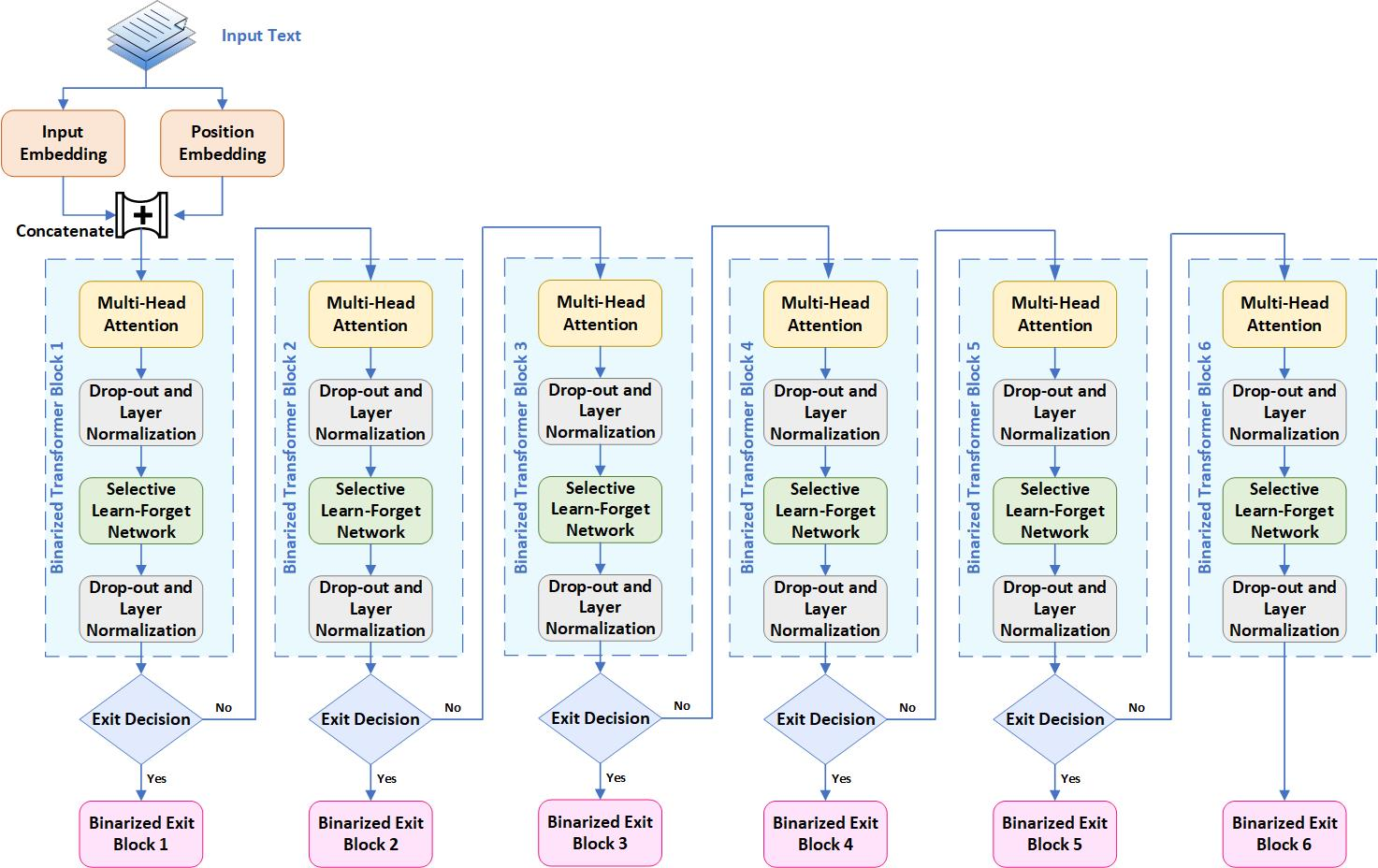}
        \caption{Overall architecture}
        \label{fig1a}       
    \end{subfigure}%
    
    \begin{subfigure}[h]{0.432\linewidth}
        \centering
        \includegraphics[width=\linewidth, height=7.2cm]{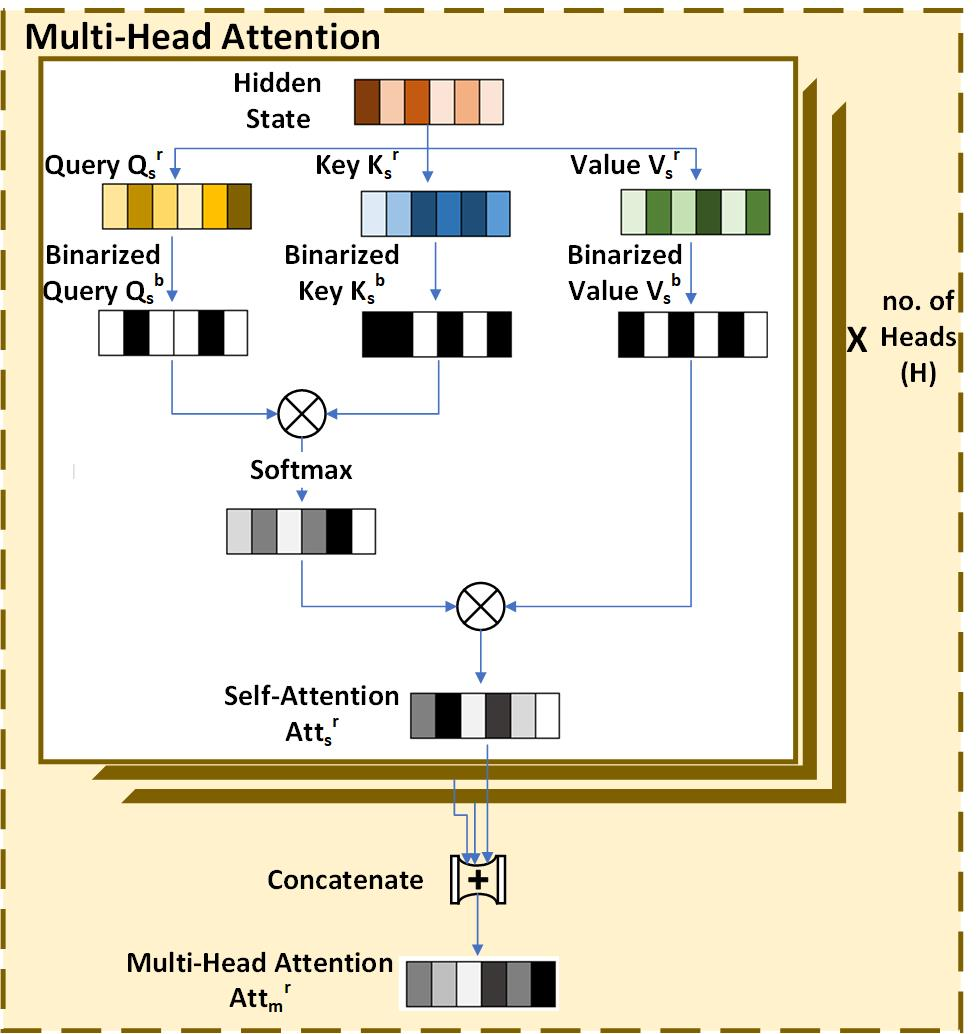}
        \caption{Binarized MHA block}
        \label{fig1b}       
    \end{subfigure}
    ~
    \begin{subfigure}[h]{0.345\linewidth}
        \centering
        \includegraphics[width=\linewidth, height=7.2cm]{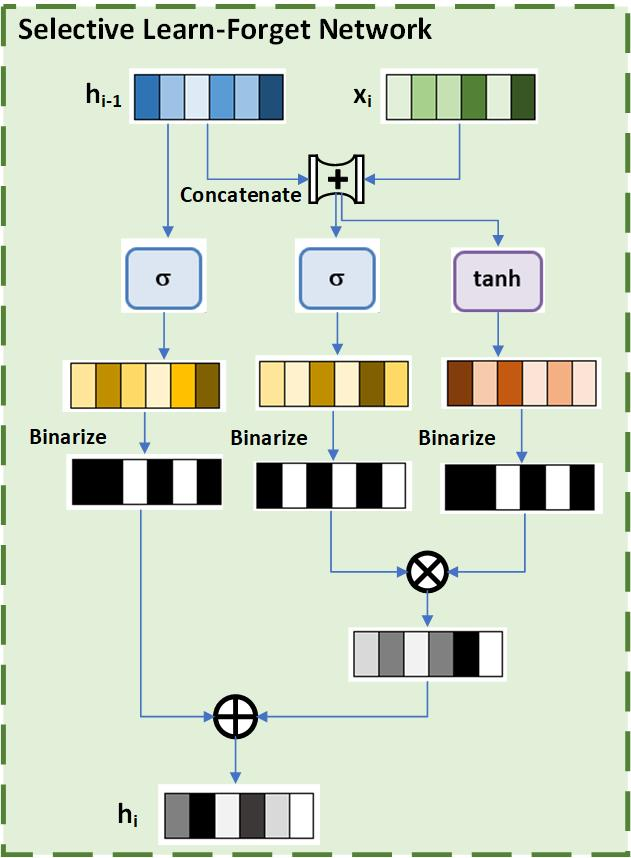}
        \caption{Binarized SLFN block}
        \label{fig1c}       
    \end{subfigure}%
    ~ 
    \begin{subfigure}[h]{0.18\linewidth}
        \centering
        \includegraphics[width=0.843\linewidth, height=7.08cm]{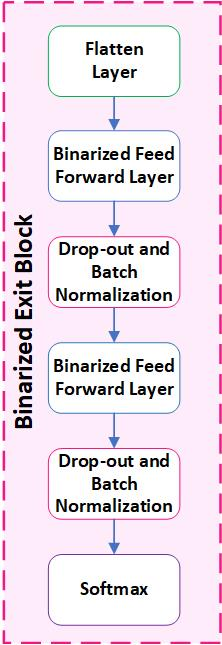}
        \caption{Binarized exit block}
        \label{fig1d}       
    \end{subfigure}
    
    \caption{Illustration of the BEExformer architecture. It comprises a binarized Multi-Head Attention (MHA) block (Figure \ref{fig1b}), a binarized Selective Learn-Forget Network (SLFN) block to handle the representation of longer contexts in an efficient manner (Figure \ref{fig1c}), and a binarized auxiliary block that executes after the exit condition is satisfied (Figure \ref{fig1d}).}
    \label{fig1}
    
\end{figure*}

The proposed BEExformer embeds an input sequence and processes it through a cascade of binarized transformer blocks incorporating SLFN interleaved with EE pathways. The proposed architecture has been illustrated in Figure \ref{fig1}, while its components have been elucidated herein-below.

\subsection{Input Representation}
\label{ir}
The input sequences undergoes pre-processing to eliminate non-desired elements such as line breaks, tabs, URLs, non-ASCII characters, and punctuation marks. Additionally, whitespace is standardized for consistency. The pre-processed sequences are tokenized, and a lookup table $LUT()$ is constituted as a one-to-one mapping between the unique tokens $V_{i}$ in the vocabulary $V$ with a unique integer $T_i$, $\forall 1 \leq i \leq |V|$. For each sentence, a tokenized sequence $S^{t}$ is constructed consulting the lookup table as shown in Algorithm \ref{alg:tsg}. Besides, padding is appended to ensure uniform sequence length $L$.

\begin{algorithm}
    \caption{Tokenized Sequence Generation}
    \label{alg:tsg}
    \begin{algorithmic}
    \Require $\textrm{Input Sequence } S^{ip} \textrm{, }$ 
            $\textrm{Vocabulary } V \textrm{, }$
            $\textrm{Lookup Table } LUT() \textrm{, }$
            $\textrm{Maximum Sequence Length } L$
    \Ensure $\textrm{Tokenized Sequence } S^{t}$

    \State $sl \gets len(S^{ip})$ 
    \Comment{Determining input sequence length}
    \For{$i \gets 1$ to $L$}
        \If{$(i \leq sl)$} 
        \Comment{For tokens in $V$}
            \State $T_i \gets LUT(V_{i})$
            \State $S^{t}_{i} \gets T_i$
        \Else 
        \Comment{For padding}
            \State $S^{t}_{i} \gets 0$
            
        \EndIf
    
    \EndFor
            
    \end{algorithmic}
    \end{algorithm}

$S^{t}$ is then converted into a series of embeddings $S^{e} \in \Re^{L \times D'} $, where each token $S^{t}_{i}$ in $S^{t}$ is substituted by an embedding vector $e[i]$. The embeddings are randomly initialized and optimized while training, as other network parameters. The position embedding $S^{p} \in \Re^{L \times D'} $ is used to determine the token ordering for contextualized representation. For a given position $i$ and dimension $d$, $S^{p}(i,d)$ is computed as shown in Equation (\ref{eq23}).

\begin{equation}
\label{eq23}
S^{p}(i,d) = \begin{cases}
          \sin{\left(\frac{i}{10000^{d/D'}}\right)} & \text{, if } d \pmod{2} = 0\\
          \cos{\left(\frac{i}{10000^{(d-1)/D'}}\right)} & \text{, if } d \pmod{2} = 1\\
        \end{cases}    
        \textrm{} 
\end{equation}

Finally, the token embedding sequence $S^{e}$ and the position-encoded sequence $S^{p}$ are element-wise concatenated (denoted by $ || $) to obtain the embedded sequence $S \in \Re^{L \times D}$ such that $D=2D'$ as shown in Equation (3).

\begin{equation}
\label{eq25}
    S= S^{e} || S^{p}
\end{equation}

\subsection{Backbone Network}
\label{oa}
The backbone network consists of $C$ transformer blocks with an exit provision after each block, as depicted in Figure \ref{fig1a}. It can be functionally represented as $F(x)$ having a cascade of $C$ functions (each $f$ refers to a transformer block presented in Section \ref{tb}) applied to the embedded input $x$ as follows:

\begin{equation}
\label{eq1}
F(x)=(f_{C} \circ f_{C-1} \circ ... \circ f_{2} \circ f_{1})(x)
\end{equation}

where $\circ$ denotes the composition operator. The hidden representation $h_{c}$ of the $c^{th}$ intermediate block is given as:

\begin{equation}
\label{eq2}
h_{c}=(f_{c} \circ f_{c-1} \circ ... \circ f_{2} \circ f_{1})(x) \textrm{, } \forall 1\leq c \leq C
\end{equation}

Alternatively, $h_{c}$ can be computed from the hidden representation $h_{c-1}$ of the $(c-1)^{th}$ block as:

\begin{equation}
\label{eq3}
\Rightarrow h_{c}=f_{c}(h_{c-1})
\end{equation}

To achieve EE, a criterion is defined. In case the criterion is fulfilled at exit $c$, $h_{c}$ is fed to its connected auxiliary block $\Phi_{c}(\cdot)$ and the probability distribution $\hat{y_{c}}$ is returned. This has been explained in Section \ref{ee}.

Additionally, BEExformer utilizes BAT (described in Section \ref{bn}) to binarize the weights and activations of all transformer blocks, as well as the auxiliary blocks for EE. Generically, a binarized layer $l$ is defined as follows:

\begin{equation}
\label{eq7}
A^{r}_{l}=a(f(b(W^{r}_{l}), b(A^{r}_{l-1}))+\beta_{l})
\end{equation}

where $A^{r}_{l}$, $a(\cdot)$, $f(\cdot)$, $b(\cdot)$, $W^{r}_{l}$, $A^{r}$ and $\beta_{l}$ denote the computed activation, activation function, layer operator, binarization function, real-valued weights, incoming activation, and bias, respectively.

\subsection{Binarization-Aware Training (BAT)}
\label{bn}

Conventionally for binarization, $sign(r)$ (Equation (\ref{eq24})) is applied, where $r$ represents a real-valued weight or activation. However, being non-differentiable, it cannot be applied to compute the gradient. To ameliorate this, we apply magnitude-aware approximation to $sign(r)$ \cite{Ref19} for the first time to the best of our knowledge in a transformer encoder. Denoted by $b(r)$, it has been expressed in Equation (\ref{eq5}). Given $b(r)$, its derivative is computed as in Equation (\ref{eq6}). $b(r)$ closely approximates $sign(r)$ equipped with the capability of being differentiable. This has been further analyzed in Section \ref{bd} through comparative plots of $b(r)$ and its derivatives alongside other binarization functions (Figure \ref{fig4}).

\begin{equation}
\label{eq24}
sign(r)=
\begin{cases}
-1 \textrm{, if } r<0  \\
1 \textrm{, if } r \geq 0 \\
\end{cases}
\end{equation}

\begin{equation}
\label{eq5}
b(r)=
\begin{cases}
-1 \textrm{, if } r \leq -1  \\
2r+r^{2} \textrm{, if } -1 < r < 0 \\
2r-r^{2} \textrm{, if } 0 \leq r < 1 \\
1 \textrm{, if } r \geq 1 \\
\end{cases}
\end{equation}

\begin{equation}
\label{eq6}
b'(r)=\frac{\partial{b(r)}}{\partial{r}}=
\begin{cases}
2(1- |r|) \textrm{, if } -1 < r < 1  \\
0 \textrm{, if } r \leq -1 \textrm{ or } r \geq 1 \\
\end{cases}
\end{equation}

where $b(r)$ is a differentiable piecewise polynomial function being a second-order approximation to the sign function $sign(r)$. While $\frac{\partial{b(r)}}{\partial{r}}$ is a piecewise linear function approximating the derivative of $sign(r)$, which is an impulse function during gradient calculation. 

Conventional gradient descent cannot be applied to BNN due to the gradient of loss $\mathcal{L}$ concerning binarized weights being insufficient to enable bit-flips. To mitigate this issue, we perform BAT wherein the gradient computed for the $l^{th}$ layer concerning binarized weights $W_{l, t}^{b}$ for the $t^{th}$ iteration is used to update real-valued weights $W_{l, t+1}^{r}$ for the $(t+1)^{th}$ iteration. This has been shown in the following equation:

\begin{equation}
\label{eq8}
W_{l, t+1}^{r}=W_{l, t}^{r} - \eta \frac{\partial{\mathcal{L}}}{\partial{W_{l, t}^{r}}}
\end{equation}

Applying the chain rule for partial derivatives on $\frac{\partial{\mathcal{L}}}{\partial{W_{l, t}^{r}}}$,

\begin{equation}
\label{eq9}
\Rightarrow W_{l, t+1}^{r}=W_{l, t}^{r} - \eta \frac{\partial{\mathcal{L}}}{\partial{W_{l,t}^{b}}}\frac{\partial{W_{l, t}^{b}}}{\partial{W_{l, t}^{r}}}
\end{equation}

\begin{equation}
\label{eq32}
\Rightarrow W_{l, t+1}^{r}=W_{l, t}^{r} - \eta \frac{\partial{\mathcal{L}}}{\partial{W_{l,t}^{b}}}\frac{\partial{b(W_{l, t}^{r})}}{\partial{W_{l, t}^{r}}}
\end{equation}

where, $W_{l, t}^{b}=b(W_{l, t}^{r})$, and $\eta$ denotes the learning rate. Applying the chain rule, $\frac{\partial{\mathcal{L}}}{\partial{W_{l,t}^{b}}}$ is expanded as follows:

\begin{equation}
\label{eq26}
\frac{\partial{\mathcal{L}}}{\partial{W_{l,t}^{b}}}= \frac{\partial{\mathcal{L}}}{\partial{A_{l,t}^{r}}}\frac{\partial{A_{l,t}^{r}}}{\partial{Z_{l, t}^{r}}}\frac{\partial{Z_{l, t}^{r}}}{\partial{W_{l,t}^{b}}}
\end{equation}

where, $A_{l,t}^{r}$ represents the real-valued activation computed for the $l^{th}$ layer and $t^{th}$ iteration, and $Z_{l, t}^{r}$ represents the output of the layer operator $f(\cdot)$ in Equation (\ref{eq7}). Since $A_{l,t}^{r}=a(Z_{l, t}^{r})$, it can be written as:

\begin{equation}
\label{eq27}
\Rightarrow \frac{\partial{\mathcal{L}}}{\partial{W_{l,t}^{b}}}= \frac{\partial{\mathcal{L}}}{\partial{A_{l,t}^{r}}}a'(Z_{l, t}^{r})b(A_{l-1,t}^{r})
\end{equation}

Expanding $\frac{\partial{\mathcal{L}}}{\partial{A_{l,t}^{r}}}$, in Equation (\ref{eq27}) applying the chain rule as follows:

\begin{equation}
\label{eq28}
\Rightarrow \frac{\partial{\mathcal{L}}}{\partial{W_{l,t}^{b}}}= \frac{\partial{\mathcal{L}}}{\partial{A_{l,t}^{b}}}\frac{\partial{A_{l,t}^{b}}}{\partial{A_{l,t}^{r}}}a'(Z_{l, t}^{r})b(A_{l-1,t}^{r})
\end{equation}

where, $A_{l,t}^{b}$, i.e. the binarized activation for the $l^{th}$ layer and $t^{th}$ iteration is computed as $b(A_{l,t}^{r})$. Therefore, Equation (\ref{eq28}) can be simplified as:

\begin{equation}
\label{eq29}
\Rightarrow \frac{\partial{\mathcal{L}}}{\partial{W_{l,t}^{b}}}= \frac{\partial{\mathcal{L}}}{\partial{A_{l,t}^{b}}}\frac{\partial{b(A_{l,t}^{r})}}{\partial{A_{l,t}^{r}}}a'(Z_{l, t}^{r})b(A_{l-1,t}^{r})
\end{equation}

Substituting $\frac{\partial{\mathcal{L}}}{\partial{W_{l,t}^{b}}}$ from Equation (\ref{eq29}) in Equation (\ref{eq32}), as follows:

\begin{equation}
\label{eq30}
W_{l, t+1}^{r}=W_{l, t}^{r} - \eta \left[\frac{\partial{\mathcal{L}}}{\partial{A_{l,t}^{b}}}\frac{\partial{b(A_{l,t}^{r})}}{\partial{A_{l,t}^{r}}}a'(Z_{l, t}^{r})b(A_{l-1,t}^{r})\right] \frac{\partial{b(W_{l, t}^{r})}}{\partial{W_{l, t}^{r}}}
\end{equation}

Based on Equation (\ref{eq6}), Equation (\ref{eq30}) can be formulated as follows:

\begin{equation}
\label{eq31}
\Rightarrow W_{l, t+1}^{r}=W_{l, t}^{r} - \eta \left[\frac{\partial{\mathcal{L}}}{\partial{A_{l,t}^{b}}}g'(A_{l,t}^{r})a'(Z_{l, t}^{r})b(A_{l-1,t}^{r})\right] b'(W_{l, t}^{r})
\end{equation}

Once BAT is completed, the updated binarized weights $W_{l, t+1}^{b}$ are frozen applying Equation (\ref{eq24}) to ensure $W_{l, t+1}^{b} \in \{-1,1\}$ as shown in Equation (\ref{eq33}). The details of the implementation of the above-mentioned binarization technique in the BEExformer transformer blocks are presented in Section \ref{tb}.

\begin{equation}
\label{eq33}
W_{l, t+1}^{b}=sign(W_{l, t}^{r})
\end{equation}

\subsection{Transformer Blocks}
\label{tb}

Each transformer block binarizes the multi-head attention (MHA) \cite{Ref17} $Att_{m}^{b}$ module as depicted in Figure \ref{fig1b}. $Att_{m}^{b}$ comprises $H$ binarized self-attention $Att_{s}^{b}$ heads, which in turn relies on queries $Q_{s}^{b}$, keys $K_{s}^{b}$, and values $V_{s}^{b}$ for computation as follows:

\begin{equation}
\label{eq12}
Att_{s}^{r}=softmax \left( \frac{Q_{s}^{b}(K_{s}^{b})^{T}}{\sqrt{D}} \right) V_{s}^{b}
\end{equation}

where $Q_{s}^{b}$, $K_{s}^{b}$, and $V_{s}^{b}$ have been obtained through binarized-linear (bi-linear) transformation (explained in Section \ref{bn}) on input $x \in \Re^{L \times D} $ where $L$ denotes the sequence length and $D$ denotes the dimension of hidden representations. The computations have been shown in equations (\ref{eq13} -\ref{eq15}).

\begin{equation}
\label{eq13}
Q_{s}^{b} = b(W_{q}^{r})b(x^{r})
\end{equation}

\begin{equation}
\label{eq14}
K_{s}^{b} = b(W_{k}^{r})b(x^{r})
\end{equation}

\begin{equation}
\label{eq15}
V_{s}^{b} = b(W_{v}^{r})b(x^{r})
\end{equation}

where $W_{q}^{r}, W_{k}^{r}, W_{v}^{r} \in \Re^{D \times D/H} $. Finally, $Att_{m}^{b}$ is calculated in Equation (\ref{eq16}) through projection of $H$ concatenated $Att_{s}^{r}$ through $W_{o}^{r} \in \Re^{D \times D}$ after binarization.

\begin{equation}
\label{eq16}
Att_{m}^{b} = b(Concatenate(Att_{s,i}^{r}))b(W_{o}^{r}) \textrm{, } \forall i \in H 
\end{equation}

To augment the efficacy of residual connection, a binarized SLFN is used to replace the feed-forward layer in the conventional transformer architecture. This is the modified version of SLFN proposed by Ansar et al. \cite{Ref15}. In SLFN, there are two forget gates, $Sh_{i}^{b}$ and $Sg_{i}^{b}$, which aid in the elimination of insignificant information in the previous hidden state and the total incoming activations, respectively. Besides, it has a selective learning gate $Tg_{i}^{b}$ for rational estimation of dependencies. All gates incorporate binarization of the incoming activations as shown in Equation (\ref{eq5}). The architecture has been illustrated in Figure \ref{fig1c}, while the formulation of the gates has been provided in the following equations.

\begin{equation}
\label{eq17}
Sh_{i}^{b}=\sigma[b(h_{i-1}^{r})b(U_{sg}^{r})]
\end{equation}

\begin{equation}
\label{eq18}
Sg_{i}^{b}=\sigma[(b(x_{i}^{r})b(W_{sg}^{r}) + (b(h_{i-1}^{r})b(U_{sg}^{r})]
\end{equation}

\begin{equation}
\label{eq19}
Tg_{i}^{b}= tanh[(b(x_{i}^{r})b(W_{tg}^{r}) + (b(h_{i-1}^{r})b(U_{tg}^{r})]
\end{equation}

where $x_{i}^{r}$ and $h_{i-1}^{r}$ denote full-precision input at $i^{th}$ step and $(i-1)^{th}$ hidden state, respectively. Whereas $W_{sg}^{r}$, $U_{sg}^{r}$, $W_{tg}^{r}$, and $U_{tg}^{r}$ denote weight matrices for inputs as well as hidden states, respectively. Here, we apply $b(\cdot)$ to all incoming weights and activations for binarization. Finally, the updated hidden state is calculated as follows:

\begin{equation}
\label{eq20}
h_{i}= b(Sg_{i}^{b})+[b(Sg_{i}^{b}) \odot b(Tg_{i^{b}})]
\end{equation}

\subsection{Early Exit}
\label{ee}

We propose an EE criterion that estimates proportionate entropy reduction over exits, i.e. fractional reduction in entropy concerning the previous exit. When this value falls below a certain threshold $\delta$, the exit criterion is satisfied. This has been portrayed in Algorithm \ref{alg:eem}. Here, the entropy of logits $S(h_{c})$ is computed as in Equations (\ref{eq10}) and (\ref{eq11}). Initially, the value of $S(h_{c})$ is taken as $\ln(m)$, which signifies the case where logits are evenly spread across $m$ classes.

\begin{equation}
\label{eq10}
S(h_{c})= -\sum_{j=1}^{m}{p(h_{c}^{j}) \ln(p(h_{c}^{j}))}
\end{equation}

\begin{equation}
\label{eq11}
\Rightarrow S_{t}(h_{c})= \ln(\sum_{j=1}^{m}{exp(h_{c}^{j})}) - \frac{\sum_{j=1}^{m}{h_{c}^{j} \cdot exp(h_{c}^{j})}}{\sum_{j=1}^{m}{exp(h_{c}^{j})}}
\end{equation}

\begin{algorithm}
    \caption{Early Exit Estimation}
    \label{alg:eem}
    \begin{algorithmic}
    \Require $\textrm{Input Sequence } x \textrm{, }$ 
            $\textrm{Number of classes } m \textrm{, }$
            $\textrm{EE threshold } \delta$
    \Ensure $\textrm{ Output at $c^{th}$ exit  } \hat{y_{c}}$

    \State $h_{0} \gets x$
    \Comment{Initialize incoming activation}
    \State $S(h_{0}) \gets \ln(m)$
    \Comment{Initialize entropy value}
    
    \For{$c \gets 1$ to $C$}     
        \State $h_{c} \gets f_{c}(h_{c-1}) $
        \Comment{Logits returned by $c^{th}$ block}
        \State Calculate $S(h_{c})$
        \If{$\frac{S(h_{c-1}) - S(h_{c})}{S(h_{c-1})} < \delta$}
            \Comment{Exit condition}
            \State $\hat{y_{c}}=\Phi_{c}(h_{c})$
            \Comment{Auxiliary block execution}
            \State \textbf{return} $\hat{y_{c}}$
        \EndIf
    \EndFor
            
    \end{algorithmic}
    \end{algorithm}    

Following this, an auxiliary block $\Phi_{c} (\cdot)$ is defined to process $h_{c}$ and predict the output $\hat{y}_{c}$ as follows:

\begin{equation}
\label{eq4}
\hat{y_{c}}=\Phi_{c}(h_{c}) \textrm{, } \ni \Phi_{c} \in \Phi
\end{equation}

Here, $\Phi_{c}(\cdot)$ belongs to the set of binarized feed-forward networks $\Phi=\{\Phi_{1}, \Phi_{2}, ... ,\Phi_{C}\}$ as shown in Figure \ref{fig1d}. It processes $h_{c}$ into a discrete probability distribution $\hat{y_{c}}$ as the output for the $c^{th}$ exit. To ensure minimal computational overhead, $\Phi_{c}(\cdot)$ has a minimal number of parameters compared to the backbone network $F(\cdot)$.

The proposed EE criterion is robust against diversity in inputs during inference, since only the fractional reduction in entropy is noted. This alleviates the limitation of previous works with absolute entropy thresholds to deal with diverse inputs. In addition, determining the absolute threshold is a cumbersome process in such models. Compared to it, setting $\delta$ is pretty straightforward based on the computational budget and the output confidence requirements. Moreover, the same value of $\delta$ can be used across multiple datasets.

\subsection{Soft-Routing Loss Function}
\label{lf}

The loss function L calculated at the $i^{th}$ exit is given by the following equation:

\begin{equation}
\label{eq21}
\mathcal{L}_{i}(\mathcal{I}; \theta)=\frac{1}{|\mathcal{I}|}\sum_{(x,y) \in \mathcal{I}}{\mathcal{H}(y, F_{i}(x;\theta))}
\end{equation}

Where $\mathcal{I}$ is the input data containing sequence-label pairs $(x,y)$, $F_{i}(x;\theta)$ is the probability distribution returned by the $i^{th}$ exit, $\theta$ denotes trainable parameters, and $\mathcal{H}$ is the cross-entropy function. The proposed model uses a soft-routing loss objective, i.e. combining loss from all exits to enhance the decision capability. This aids in an optimal assessment of the exit based on the predictions from all the exit blocks while reducing the loss. The total loss $\mathcal{L'}$ is calculated as the mean loss across all exits given by:

\begin{equation}
\label{eq22}
\mathcal{L'}(\mathcal{I}; \theta)=\frac{1}{|C|}\sum_{i}^{C}{\mathcal{L}_{i}(\mathcal{I}; \theta)}
\end{equation}

where C is the total number of exits.

\section{Experimental Setup}
\label{es}
In this section, details of the experiment conducted along with dataset information have been provided.

\subsection{Datasets and Metrics}
\label{ds}
The proposed methodology has been applied to nine datasets from the GLUE benchmark \cite{Ref7}, including Stanford Sentiment Treebank (SST-2), Corpus of Linguistic Acceptability (CoLA), the Microsoft Research Paraphrase Corpus (MRPC), Recognizing Textual Entailment (RTE), Multi-Genre Natural Language Inference (MNLI) having both matched (m) and mismatched (mm) variants, Semantic Textual Similarity Benchmark (STS-B), Question Natural Language Inference (QNLI), and Quora Question Pairs (QQP). These datasets cover a wide range of tasks such as sentiment analysis, linguistic acceptability, semantic similarity between texts, paraphrase detection, and entailment detection. The metrics for evaluation are F1 score for MRPC, Spearman correlation for STS-B, Matthews correlation for CoLA, and accuracy for the remaining datasets.

As metrics to measure efficiency, size of the model is reported along with GFLOPs for a holistic estimation of the storage requirements and inference time. The bit-wise operations (XNOR/ pop-count) in the BNN are estimated as GFLOPs in a similar manner to prior works by Liu et al. \cite{Ref19} and Bai et al. \cite{Ref1}. Given that modern processors can execute XNOR and pop-count in parallel across 64 bits, the GFLOPs are calculated as $1/64 \times$ the number of bit-wise operations. This is then added with the GFLOPs for the full-precision operations to obtain the total GFLOPs.

\begin{table}[t!]
  \begin{center}
    \caption{Hyperparameter Details of the BEExformer Model}
    \label{table:table2}
    \setlength{\tabcolsep}{0.5em} 
    \renewcommand{\arraystretch}{1} 
 \begin{tabular}{|l|p{5.1cm}|} 
 \hline
 Hyperparameter & Value\\ 
 \hline
 \# Transformer Blocks & 6 \\
 \# Attention Heads & 4 \\
 Embedding Dimension & 512 \\
 Hidden Layer Dimension & 768 \\
 EE Threshold ($\delta$) & 0.0001 \\
 Dropout Ratio & 0.3 \\
 Optimizer & Adam \\
 Batch Size & 32 \\
 Training Epochs & 100 \\
 Loss Function & Sparse Categorical Cross-Entropy \\
 Learning Rate (LR) & [0.01 - 0.0001] (Reduces by a factor of 0.1 upon loss plateau)  \\
 LR Loss Plateau Patience & 3 Epochs \\
 Warmup LR & [0.00001 - 0.01] \\
 Warmup Steps & 500 \\
 Warmup Scheduler Type & Linear \\
 Early Stopping Patience & 5 Epochs \\

 \hline
 
\end{tabular}
\end{center}
\end{table}

\subsection{Implementation Details}
\label{hd}

The proposed BEExformer has been implemented on a Python 3 Compute Engine with the Nvidia L4 GPU, 22.5GB VRAM, 53GB System RAM, and 201.2GB Disk in Google Colab\footnote{https://colab.research.google.com}. The details of the hyperparameters of the BEExformer have been presented in Table \ref{table:table2}.

\section{Results and Discussion}
\label{rd}

\subsection{Comparison with Related Works}
\label{cr}
To the best of our knowledge, the combination of binarization with EE is unprecedented for textual content. Thus, we compare with works which either implement QNN or EE in separate tables for ease of comprehension. In addition, we provide a comparison with standard full-precision models with an equal number of transformer blocks to analyze the impact of BAT, EE, and SLFN on model performance and efficiency.

\subsubsection{Purely Quantized Models}
\label{pqm}
Table \ref{table:table1} presents the comparison with SOTA Quantized Models without EE on the basis of model quantization configuration, size, GFLOPs, and performance metrics for various datasets. BEExformer stands out with a remarkably low 0.12 GFLOPs, compared to quantized models without EE, demonstrating the efficiency of EE in reducing computation during inference. By dynamically eliminating processing through a certain number of layers, it avoids full deep-network computation while maintaining a compact 14.26 MB model size, comparable to BiBERT and BiT, and even smaller than BinaryBERT—indicating minimal overhead from EE and SLFN. Despite not distilling knowledge from full-precision models like BERT, BEExformer achieves strong benchmark performance, outperforming other BNNs with (1-1) quantization configuration. Among other models, binarized versions of BinaryBERT, BiBERT and BiT offer small size and low GFLOPs, their performance drops notably on CoLA and RTE, implying that binarization alone is insufficient without mechanisms like EE or SLFN. In contrast, TernaryBERT and Q-BERT achieve competitive performance but at the cost of larger size and higher GFLOPs, demonstrating a suboptimal trade-off between performance and efficiency. Interestingly, BiBERT-6L with 6 layers, experiences just 1.1\% drop in overall performance compared to the default 12 layer BiBERT while surpassing it on STS-B and QNLI. This indicates that additional blocks beyond a certain extent does not contribute much to performance, strengthening our EE proposition. Furthermore, BEExformer delivers superior performance than BiBERT-6L. Despite being larger in size due to the additional EE and SLFN mechanisms, it takes 40\% lower GFLOPs due to EE-based dynamic inference. Overall, BEExformer provides an optimal balance due to integration with SLFN and EE, providing high performance with superior efficiency.

\begin{table*}[h!]
  \begin{center}
    \caption{Comparison with Quantized Models without EE}
    \label{table:table1}
    \setlength{\tabcolsep}{0.5em} 
    \renewcommand{\arraystretch}{1} 
 \resizebox{\textwidth}{!}{\begin{tabular}{|p{3cm}|p{1cm}|p{1cm}|p{1.2cm}|p{0.72cm}|p{0.66cm}|p{0.72cm}|p{0.6cm}|p{0.81cm}|p{0.81cm}|p{0.72cm}|p{0.72cm}|p{0.72cm}|p{0.72cm}|} 
 \hline
 Model & \#Bits $\downarrow$ (W-A) & Size $\downarrow$ (MB) & GFLOPs $\downarrow$ & SST-2 $\uparrow$ & CoLA $\uparrow$ & MRPC $\uparrow$ & RTE $\uparrow$ & MNLI-m $\uparrow$ & MNLI-mm $\uparrow$ & STS-B $\uparrow$ & QNLI $\uparrow$ & QQP $\uparrow$ & Avg. $\uparrow$\\
 \hline
 \multicolumn{14}{|c|}{\textbf{Quantized Models without EE}} \\
 \hline
 Q-BERT \cite{Ref12} & 8-8 & 43 & 6.5 & 84.6 & - & 68.3 & 52.7 & 76.6 & 77.0 & - & - & - & 71.8 \\
 TernaryBERT \cite{Ref13} & 2-8 & 28 & 6.4 & - & 50.7 & \textbf{87.5} & 68.2 & 83.3 & 83.3 & - & - & 90.1 & 77.2 \\
 TernaryBERT \cite{Ref13} & 2-2 & 28 & 1.5 & 80.7 & - & 68.3 & 54.5 & 40.3 & 40.0 & 12.4 & 50.0 & 63.1 & 51.2 \\ 
 TernaryBERT \cite{Ref13} & 2-1 & 28 & 0.8 & 53.1 & - & 68.3 & 53.4 & 32.7 & 33.0 & 7.1 & 59.3 & 74.1 & 47.6 \\
 BinaryBERT \cite{Ref1} & 1-8 & 16.5 & 3.1 & \textbf{92.6} & \textbf{53.4} & 85.5 & \textbf{72.2} & \textbf{84.2} & \textbf{84.7} & \textbf{88.6} & \textbf{91.5} & \textbf{91.2} & \textbf{82.7} \\
 BinaryBERT \cite{Ref1} & 1-4 & 16.5 & 1.5 & 92.3 & 44.4 & 83.3 & 65.3 & 83.9 & 84.2 & 87.2 & 90.9 & \textbf{91.2} & 80.3 \\ 
 BinaryBERT \cite{Ref1} & 1-2 & 16.5 & 0.8 & 82.5 & 14.6 & 68.3 & 52.7 & 62.7 & 63.9 & 6.5 & 52.6 & 79.9 & 53.7 \\ 
 BinaryBERT \cite{Ref1} & 1-1 & 16.5 & 0.4 & 53.2 & 0 & 68.3 & 52.7 & 35.6 & 35.3 & 6.1 & 51.5 & 66.2 & 41.0 \\ 
 BiBERT  \cite{Ref2} & 1-1 & 13.4 & 0.4 & 88.7 & 25.4 & 72.5 & 57.4 & 66.1 & 67.5 & 33.6 & 72.6 & 84.8 & 63.2 \\ 
 BiBERT-6L \cite{Ref2} & 1-1 & \textbf{6.8} & \textbf{0.2} & 87.9 & 24.8 & 72.2 & 55.9 & 63.6 & 63.7 & 33.7 & 73.6 & 83.3 & 62.1 \\ 
 BiT (WMS) \cite{Ref3} & 1-4 & 13.4 & 1.5 & 91.5 & 42.0 & 86.8 & 66.4 & 83.6 & 84.4 & 86.3 & 91.3 & 87.8 & 80.0 \\ 
 BiT (WMS) \cite{Ref3} & 1-2 & 13.4 & 0.8 & 90.8 & 32.1 & 78.4 & 58.1 & 82.1 & 82.5 & 82.2 & 89.3 & 87.1 & 75.8 \\
 BiT (WMS) \cite{Ref3} & 1-1 & 13.4 & 0.4 & 87.7 & 25.1 & 79.7 & 58.8 & 77.1 & 77.5 & 71.1 & 85.7 & 82.9 & 71.7 \\
 BiT \cite{Ref3} & 1-1 & 13.4 & 0.4 & 89.9 & 32.9 & 79.9 & 62.1 & 79.5 & 79.4 & 72.0 & 86.4 & 85.4 & 74.2 \\
 \hline
 \multicolumn{14}{|c|}{\textbf{Binarization with EE}} \\
 \hline
 BEExformer (Proposed) & 1-1 & \textbf{14.26} & \textbf{0.12}  & \textbf{92.32} & \textbf{60.30} & \textbf{86.47} & \textbf{71.11} & \textbf{83.71} & \textbf{84.29} & \textbf{86.60} & \textbf{89.18} & \textbf{88.09} & \textbf{82.45} \\
 \hline
 \multicolumn{14}{p{\linewidth}}{Note- WMS: Without multi-distillation, Avg.: Average score across all datasets, '-' indicates value is not available. \newline 
 The suffix '$-\lambda L$' has been appended, where $\lambda$ represents the number of layers (transformer blocks). \newline
 The default metrics for each dataset have been used for the results. \newline
 $\uparrow$ indicates higher value is desired while $\downarrow$ indicates lower value is desired.
The boldfaced values indicate the best results across each category.} \\
 
\end{tabular}}
\end{center}
\end{table*}

\begin{table*}[h!]
  \begin{center}
    \caption{Comparison with Full-precision EE Models}
    \label{table:table5}
    \setlength{\tabcolsep}{0.5em} 
    \renewcommand{\arraystretch}{1} 
 \resizebox{\textwidth}{!}{\begin{tabular}{|p{3.6cm}|p{1cm}|p{1cm}|p{1.2cm}|p{0.72cm}|p{0.66cm}|p{0.72cm}|p{0.6cm}|p{0.81cm}|p{0.81cm}|p{0.72cm}|p{0.72cm}|p{0.72cm}|p{0.72cm}|}
 \hline
 Model & \#Bits $\downarrow$ (W-A) & Size $\downarrow$ (MB) & GFLOPs $\downarrow$ & SST-2 $\uparrow$ & CoLA $\uparrow$ & MRPC $\uparrow$ & RTE $\uparrow$ & MNLI-m $\uparrow$ & MNLI-mm $\uparrow$ & STS-B $\uparrow$ & QNLI $\uparrow$ & QQP $\uparrow$ & Avg. $\uparrow$ \\
 \hline

 \hline
 \multicolumn{14}{|c|}{\textbf{Full-precision EE Models}} \\
 \hline
 LayerDrop\textsubscript{Base} \cite{fan2019reducing} & 32-32 & 477.5 & 6.7 & 93.7 & 64.5 & 91.6 & 71.1 & 86.4 & 86.5 & 88.6 &  92.2 &  89.9 &  84.9 \\
 LayerDrop\textsubscript{Base}-6L \cite{fan2019reducing} & 32-32 & 313.2 & 3.4 &  91.3 & 53.7 & 87.6 & 64.3 & 83.8 & 83.8 & 88.1 &  89.8 &  89.4 &  81.3 \\
 LayerDrop\textsubscript{Large} \cite{fan2019reducing} & 32-32 & 1356.1 & 23.8 & 95.5 & 66.6 & 91.2 & \textbf{86.6} & \textbf{89.7} & \textbf{89.6} & \textbf{92.6} &  \textbf{93.9} &  88.5 &  88.2 \\
 LayerDrop\textsubscript{Large}-6L \cite{fan2019reducing} & 32-32 & 492.8 & 6.0 &  91.4 & 44.3 & 79.8 & 53.1 & 81.4 & 81.0 & 83.0 &  87.1 &  88.5 &  76.6 \\
 HeadPruneBERT\textsubscript{Base} \cite{michel2019sixteen} & 32-32 & 332.3 & 4.7 & 89.4 & 48.7 & 80.2 & 62.5 & 71.0 & 79.7 & 85.2 &  86.1 &  84.7 &  76.4 \\
 ElasticBERT\textsubscript{Base} \cite{Ref14} & 32-32 & 416.4 & 6.6 & 94.3 & 64.3 & 91.0 & 76.5 & 85.3 & 85.9 & 90.7 &  92.0 &  90.2 &  85.6 \\
 ElasticBERT\textsubscript{Base}-6L \cite{Ref14} & 32-32 & 255.9 & \textbf{3.3} &  92.7 & 53.7 & 89.7 & 74.0 & 84.3 & 84.2 & 90.2 &  90.8 &  89.7 &  83.3 \\
 ElasticBERT\textsubscript{Large} \cite{Ref14} & 32-32 & 1279.7 & 23.4 & 95.3 & 66.3 & \textbf{92.0} & 83.1 & 88 & 88.5 & 91.7 &  93.6 &  \textbf{90.9} &  87.7 \\
 ElasticBERT\textsubscript{Large}-6L \cite{Ref14} & 32-32 & 412.6 & 5.9 &  91.9 & 53.9 & 89.6 & 71.1 & 83.5 & 84.3 & 90.1 &  90.8 &  90.1 &  82.8 \\
 BE3R\_BERT \cite{Ref6} & 32-32 & 1297 & - & 93.2 & 59.8 & 90.3 & 71.5 & 85.6 & 85.9 & 89.4 &  91.9 &  88.1 &  84.0 \\
 BE3R\_Electra \cite{Ref6} & 32-32 & 1278 & - & \textbf{96.8} & \textbf{68.3} & 91.5 & 86.3 &  89.6 & 89.3 & 92.3 &  93.7 &  89.6 &  \textbf{88.6} \\
 PABEE \cite{Ref4} & 32-32 & \textbf{46} & 4.4 & 93.0 & 61.2 & 90.0 & 80.1 & 85.1 & 85.1 & 90.1 &  91.8 &  89.6 &  85.1 \\
 \hline
 \multicolumn{14}{|c|}{\textbf{Binarization with EE}} \\
 \hline
 BEExformer (Proposed) & 1-1 & \textbf{14.26} & \textbf{0.12}  & \textbf{92.32} & \textbf{60.30} & \textbf{86.47} & \textbf{71.11} & \textbf{83.71} & \textbf{84.29} & \textbf{86.60} & \textbf{89.18} & \textbf{88.09} & \textbf{82.45} \\
 \hline
 \multicolumn{14}{p{\linewidth}}{Note- Avg.: Average score across all datasets, '-' indicates value is not available.\newline
 The suffix '$-\lambda L$' has been appended, where $\lambda$ represents the number of layers (transformer blocks). \newline The default metrics for each dataset have been used for the results. \newline
 $\uparrow$ indicates higher value is desired while $\downarrow$ indicates lower value is desired.
The boldfaced values indicate the best results across each category.} \\
 
\end{tabular}}
\end{center}
\end{table*}

\begin{table*}[h!]
  \begin{center}
    \caption{Comparison with Full-precision Standard Transformer Models}
    \label{table:table6}
    \setlength{\tabcolsep}{0.5em} 
    \renewcommand{\arraystretch}{1} 
 \resizebox{\textwidth}{!}{\begin{tabular}{|p{3.6cm}|p{1cm}|p{1cm}|p{1.2cm}|p{0.72cm}|p{0.66cm}|p{0.72cm}|p{0.6cm}|p{0.81cm}|p{0.81cm}|p{0.72cm}|p{0.72cm}|p{0.72cm}|p{0.72cm}|}
 \hline
 Model & \#Bits $\downarrow$ (W-A) & Size $\downarrow$ (MB) & GFLOPs $\downarrow$ & SST-2 $\uparrow$ & CoLA $\uparrow$ & MRPC $\uparrow$ & RTE $\uparrow$ & MNLI-m $\uparrow$ & MNLI-mm $\uparrow$ & STS-B $\uparrow$ & QNLI $\uparrow$ & QQP $\uparrow$ & Avg. $\uparrow$ \\
 \hline

 \hline
 \multicolumn{14}{|c|}{\textbf{Full-precision Standard Transformer Models Having Default Architecture}} \\
 \hline
 BERT\textsubscript{Base}-12L \cite{devlin2019bert} & 32-32 & 418 & \textbf{6.62} & 92.9 & 56.5 & 87.6 & 69.0 & 84.6 & 84.9 & 89.4 & 91.2 & 89.6 & 82.9 \\
 ALBERT\textsubscript{Base}-12L \cite{lan2019albert} & 32-32 & \textbf{46} & 6.86 & 92.8 & 56.8 & 90.5 & 78.3 & 84.9 & 85.6 &  90.7 & 91.4 & 89.2 & 84.5 \\
 RoBERTa\textsubscript{Base}-12L \cite{liu2019roberta} & 32-32 & 479 & 6.73 & 94.8 & 63.6 & 90.8 & 77.5 & 87.5 & 87.2 &  90.9 & 92.7 & 90.3 & 86.1 \\
 BERT\textsubscript{Large}-24L \cite{devlin2019bert} & 32-32 & 1285 & 23.45 & 93.5 & 61.6 & 90.1 & 72.9 & 86.2 & 86.0 &  90.4 & 92.2 & 90.1 & 84.8 \\
 ALBERT\textsubscript{Large}-24L \cite{lan2019albert} & 32-32 & 69 & 24.30 & 95.2 & 60.1 & 90.4 & 83.0 & 86.0 & 86.1 &  91.4 & 91.6 & 89.6 & 85.9 \\
 RoBERTa\textsubscript{Large}-24L \cite{liu2019roberta} & 32-32 & 1362 & 23.84 & \textbf{95.4} & \textbf{66.4} & \textbf{91.6} & \textbf{86.6} & \textbf{89.0} & \textbf{89.6} &  \textbf{92.3} & \textbf{94.2} & \textbf{90.7} & \textbf{88.4} \\
 
 \hline
 \multicolumn{14}{|c|}{\textbf{Full-precision Standard Transformer Models Having 6-Layers}} \\
 \hline
 BERT\textsubscript{Base}-6L \cite{devlin2019bert} & 32-32 & 257 & \textbf{3.31} & 90.9 & 44.6 & 84.9 & 65.7 & 81.4 & 81.4 & 88.1 & 87.4 & 88.7 & 79.2 \\
 ALBERT\textsubscript{Base}-6L \cite{lan2019albert} & 32-32 & \textbf{46} & 3.44 & 90.8 & \textbf{52.4} & \textbf{89.0} & \textbf{70.4} & 82.6 & 82.2 & \textbf{89.6} & 89.8 & 88.7 & \textbf{81.7} \\
 RoBERTa\textsubscript{Base}-6L \cite{liu2019roberta} & 32-32 & 314 & 3.36 & 92.1 & 44.4 & 87.9 & 60.6 & \textbf{84.2} & \textbf{84.6} & 89.0 & \textbf{90.5} & \textbf{89.8} & 80.3 \\
 BERT\textsubscript{Large}-6L \cite{devlin2019bert} & 32-32 & 414 & 5.86 & 89.7 & 20.2 & 76.4 & 58.5 & 76.5 & 76.5 & 77.3 & 84.3 & 87.3 & 71.9 \\
 ALBERT\textsubscript{Large}-6L \cite{lan2019albert} & 32-32 & 69 & 6.08 & \textbf{92.2} & 51.7 & 86.5 & 66.4 & 82.2 & 82.9 & 89.4 & 89.4 & 88.6 & 81.0 \\
 RoBERTa\textsubscript{Large}-6L \cite{liu2019roberta} & 32-32 & 495 & 5.96 & 90.1 & 43.3 & 80.0 & 54.9 & 80.4 & 80.9 & 80.5 & 86.1 & 88.9 & 76.1 \\
 \hline
 \multicolumn{14}{|c|}{\textbf{Proposed Model}} \\
 \hline
 BEExformer (Proposed) & 1-1 & \textbf{14.26} & \textbf{0.12}  & \textbf{92.32} & \textbf{60.30} & \textbf{86.47} & \textbf{71.11} & \textbf{83.71} & \textbf{84.29} & \textbf{86.60} & \textbf{89.18} & \textbf{88.09} & \textbf{82.45} \\
 \hline
 \multicolumn{14}{p{\linewidth}}{Note- Avg.: Average score across all datasets.\newline
 The suffix '$-\lambda L$' has been appended, where $\lambda$ represents the number of layers (transformer blocks). \newline The results of the model variants are from Liu et al. \cite{Ref14}.
 The default metrics for each dataset have been used for the results. \newline
 $\uparrow$ indicates higher value is desired while $\downarrow$ indicates lower value is desired.
The boldfaced values indicate the best results across each category.} \\
 
\end{tabular}}
\end{center}
\end{table*}

\subsubsection{Purely EE Models}
\label{pem}
In Table \ref{table:table5} comparing purely EE models, BEExformer stands out for its efficiency, having 14.26 MB size, i.e. 46 $\times$ smaller than the average of listed models, and requiring just 0.12 GFLOPs for inference. In contrast, full-precision models span hundreds to more than a thousand MB, demanding 3.31–24.30 GFLOPs. Although it excels in efficiency, it exhibits slightly less performance compared to full-precision SOTA models such as LayerDropLarge and BE3R\_Electra. However, BEExformer remains highly competitive, achieving performance comparable to most models and even outperforming some like HeadPruneBERT\textsubscript{Base} on various datasets. Comparing the 6-layer variants, the average performance of BEExformer surpasses LayerDrop\textsubscript{Base}-6L and LayerDrop\textsubscript{Large}-6L while slightly lagging behind ElasticBERT\textsubscript{Base}-6L and ElasticBERT\textsubscript{Large}-6L. Therefore, BEExformer excels in balancing efficiency and performance, delivering high competitiveness with full-precision EE models.

\begin{table*}[h!]
  \begin{center}
    \caption{Comparison of Model Ablations}
    \label{table:table4}
    \setlength{\tabcolsep}{0.5em} 
    \renewcommand{\arraystretch}{1} 
 \resizebox{\textwidth}{!}{\begin{tabular}{|p{4.8cm}|p{1cm}|p{1cm}|p{0.72cm}|p{0.66cm}|p{0.72cm}|p{0.6cm}|p{0.81cm}|p{0.81cm}|p{0.72cm}|p{0.72cm}|p{0.72cm}|p{0.72cm}|}
 \hline
 Model & \#Bits $\downarrow$ (W-A) & Size $\downarrow$ (MB)  & SST-2 $\uparrow$ & CoLA $\uparrow$ & MRPC $\uparrow$ & RTE $\uparrow$ & MNLI-m $\uparrow$ & MNLI-mm $\uparrow$ & STS-B $\uparrow$ & QNLI $\uparrow$ & QQP $\uparrow$ & Avg. $\uparrow$ \\
 \hline
 \multicolumn{13}{|c|}{\textbf{Full-Precision Ablations}} \\
 \hline
 BEExformer (WSLFN-WEE-FP) & 32-32 & \textbf{197.78} & 88.96 & 50.81 & 78.12 & 59.42 & 72.41 & 73.84 & 76.20 & 79.77 & 81.03 & 73.40 \\
 BEExformer (WSLFN-FP) & 32-32 & 238.45 & 90.71 & 57.43 & 84.27 & 67.15 & 79.39 & 80.88 & 84.13 & 85.78 & 84.36 & 79.34 \\
 BEExformer (WEE-FP) & 32-32 & 263.09 & 92.72 & 61.33 & 88.41 & 72.67 & 83.91 & 84.70 &  87.27 & 90.35 & 88.71 & 83.34 \\
 BEExformer (FP) & 32-32 & 303.76 & \textbf{95.10} & \textbf{62.70} & \textbf{90.92} & \textbf{75.81} & \textbf{87.18} & \textbf{88.09} &  \textbf{91.07} & \textbf{92.68} & \textbf{90.05} & \textbf{85.96} \\
 \hline
 \multicolumn{13}{|c|}{\textbf{Binarized Ablations}} \\
 \hline
 BEExformer (WSLFN-WEE) & 1-1 & \textbf{7.12} & 81.19 & 44.01 & 76.29 & 57.61 & 67.81 & 68.79 & 72.60 & 72.87 & 72.39 & 68.17 \\
 BEExformer (WSLFN) &  1-1 & 10.33 & 85.77 & 53.21 & 82.30 & 64.98 & 76.48 & 77.22 & 80.53 & 81.51 & 79.41 & 75.71 \\
 BEExformer (WEE) & 1-1 & 11.05 & 89.61 & 58.62 & 84.81 & 68.12 & 79.27 & 80.41 & 83.47 & 85.17 & 83.62 & 79.23 \\
 BEExformer (Proposed) & 1-1 & 14.26 & \textbf{92.32} & \textbf{60.30} & \textbf{86.47} & \textbf{71.11} & \textbf{83.71} & \textbf{84.29} & \textbf{86.60} & \textbf{89.18} & \textbf{88.09} & \textbf{82.45} \\
 \hline
 \multicolumn{13}{p{\linewidth}}{Note- WEE: Without EE ; WEE-FP: Full-Precision without EE ; FP: Full-Precision ; WSLFN: Without SLFN ; WSLFN-FP: Full-Precision without SLFN, WSLFN-WEE: Without SLFN and EE, WSLFN-WEE-FP: Full-Precision without SLFN and EE, Avg.: Average score across all datasets. The default metrics for each dataset have been used for the results. \newline
 $\uparrow$ indicates higher value is desired and $\downarrow$ indicates lower value is desired.
The boldfaced values indicate the best results across each category.} \\
\end{tabular}}
\end{center}
\end{table*}

\subsubsection{Standard Transformer Models}
\label{stm}
Table \ref{table:table6} compares BEExformer with full-precision standard transformer models based on the BERT architecture. It also includes 6-layer variants of these models to evaluate performance at a similar scale as the BEExformer. The BEExformer is 3.22 $\times$ smaller in size and requires 27.58 $\times$ fewer GFLOPs compared to the most optimal full-precision Transformer model listed in the table. Despite its efficiency, BEExformer achieves performance comparable to that of full-precision models with their default number of layers. Moreover, it outperforms 6-layer full-precision models in overall performance and specifically on SST-2, CoLA, and RTE datasets. This highlights that the combination of SLFN, BAT and EE in the proposed BEExformer makes it a computationally efficient alternative to conventional transformer models.

\subsection{Study of Model Ablations}
\label{as}

Table \ref{table:table4} presents the results from various ablations of the proposed BEExformer. Here, BEExformer without SLFN and WEE (WSLFN-WEE) is the most lightweight version, while the full-precision version (FP), gives the best results among the ablations. Despite being 21.30 $\times$ smaller than the FP ablated version, the proposed BEExformer undergoes a modest performance drop of 3.51\% across datasets. While, the proposed BEExformer delivers performance comparable to full-precision ablation without EE (WEE-FP) with a difference of 0.89\% despite being 18.45 $\times$ smaller in size. Moreover, BEExformer excels over the binarized ablation without EE (WEE) by 3.22\%. This highlights the potency of EE to boost performance by countering ``overthinking". From the SLFN ablations, it can be perceived that BEExformer leads over the full-precision ablation (WSLFN-FP), and binarized ablation (WSLFN) by 3.11\% and 6.74\% respectively. Thus SLFN plays a significant role in augmenting the transformer's capability to comprehend dependencies in the input sequence. Furthermore, BEExformer is compared with both binarized and FP ablations devoid of both SLFN and EE, i.e. (WSLFN-WEE) and (WSLFN-WEE-FP) respectively. This enables assessment of performance relative to a vanilla transformer architecture with 6 layers, and the proposed BAT mechanism. It can be observed that the performance of BEExformer (WSLFN-WEE-FP) falls behind most of the 6-layer, full-precision pre-trained variants of BERT listed in Table \ref{table:table6}. This indicates that pre-training might be effective for standard full-precision transformer-based models. However, integrating SLFN (WEE-FP) and EE (WSLFN-FP) individually makes it potent enough to deliver performance on par with the 6-layer BERT variants. Furthermore, when both SLFN and EE are included together as in BEExformer (FP), it surpasses all the 6-layer variants of BERT. Besides, the (WSLFN-WEE) and (WSLFN-WEE-FP) ablations lag behind BEExformer by 14.28\% and 9.05\% respectively in overall performance. Thus, considering the performance-efficiency trade-off, BEExformer delivers superior results over all its ablations, as detailed in Section \ref{po}.

\subsection{Pareto Optimality}
\label{po}
\begin{figure*}[h!]
    \centering
    \begin{subfigure}[h]{0.5\textwidth}
        \centering
        \includegraphics[width=0.9\linewidth, height=5.4cm]{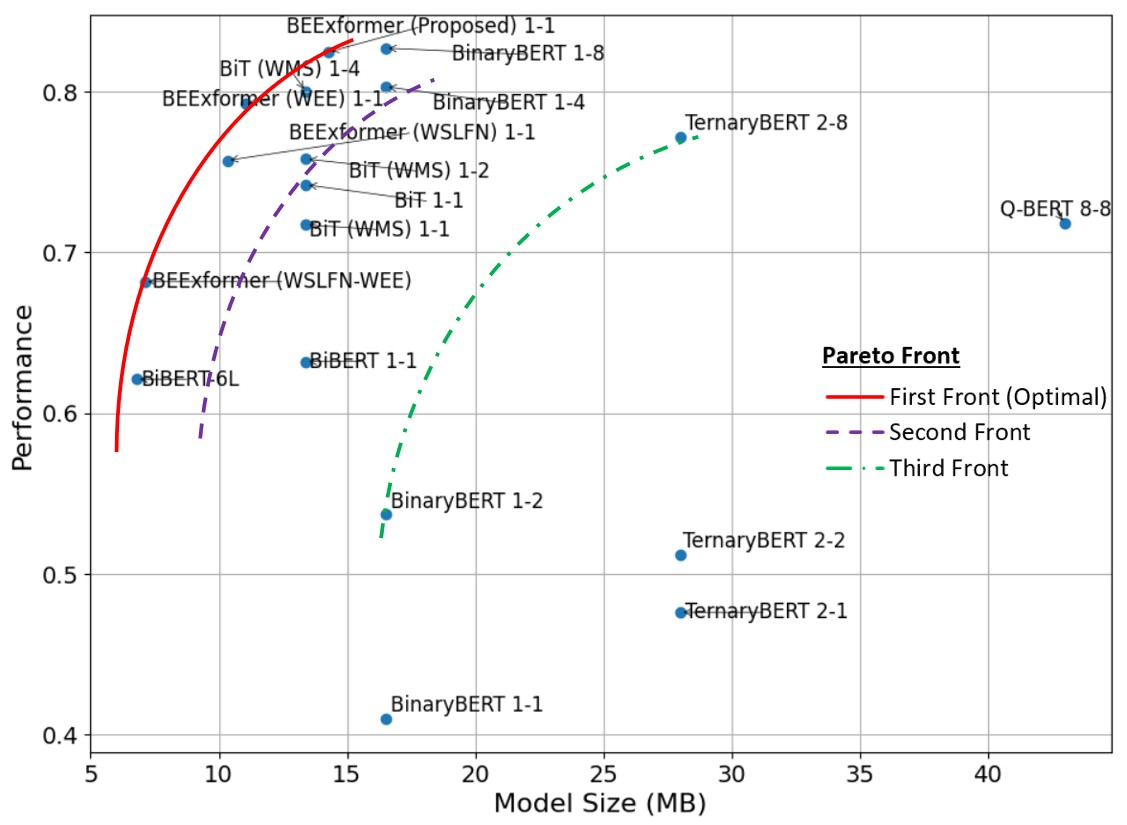}
        \caption{Comparison with quantized models}
        \label{fig3a}       
    \end{subfigure}%
    ~
    \begin{subfigure}[h]{0.5\textwidth}
        \centering
        \includegraphics[width=0.9\linewidth, height=5.4cm]{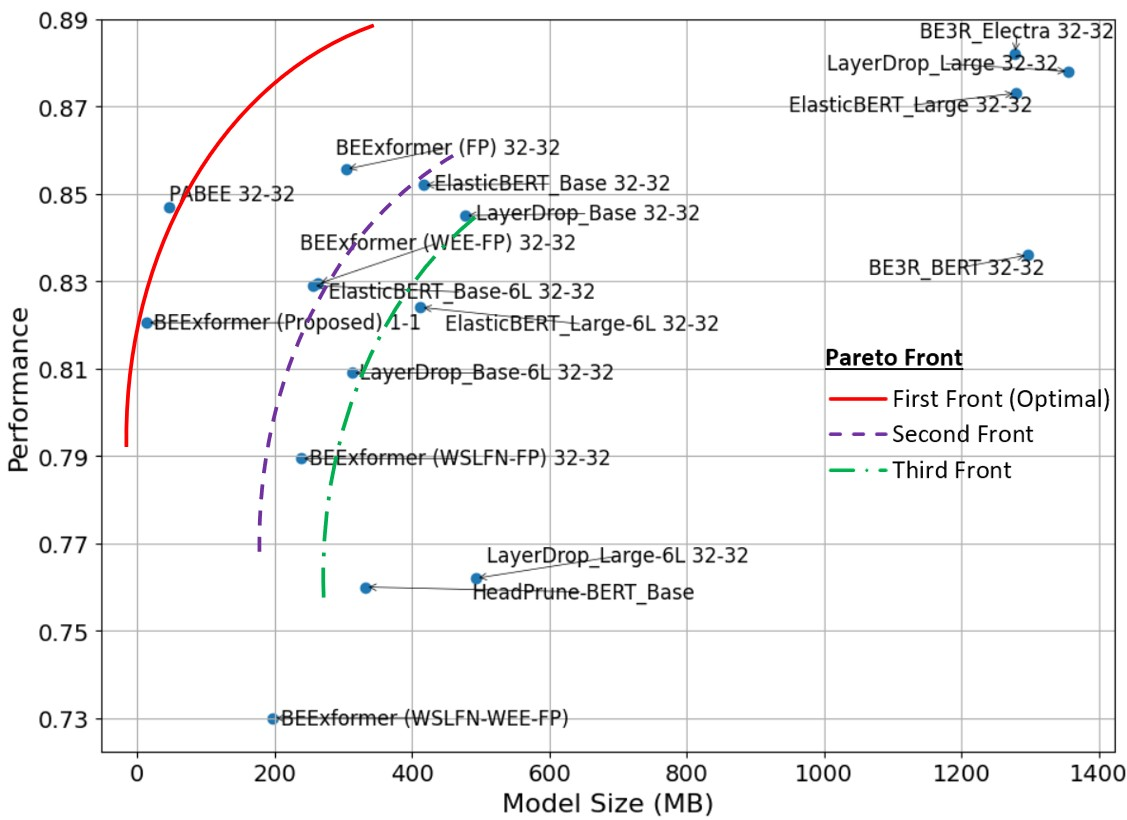}
        \caption{Comparison with EE models}
        \label{fig3b}       
    \end{subfigure}

    \caption{Pareto front charts comparing BEExformer with related works as well as its ablations. For a justified comparison, the quantized models and the EE models have been compared separately.}
    \label{fig3}
    
\end{figure*}

To evaluate the multi-objective optimization of performance along with model size, BEExformer has been compared on Pareto fronts against SOTA quantized models and EE models, as shown in Figures \ref{fig3a} and \ref{fig3b} respectively. The plots illustrate the optimal Pareto Front (solid red line), with the second and third fronts marked by dashed purple and green lines. Each point represents a model, labeled with its name and quantization configuration. Models near the optimal front achieve superior performance for a given model size or smaller sizes for a given performance level.

In Figure \ref{fig3a}, BEExformer (Proposed) 1-1 is positioned on the optimal Pareto Front, attaining the highest overall performance (82.45\%) with a compact 14.26MB size. Ablations like BEExformer (WSLFN-WEE) 1-1 and BEExformer (WEE) 1-1 remain Pareto-optimal but the performance drops by 14.28\% and 3.22\% for a model size reduction by 7.14MB and 3.21MB, respectively. Thus, removing WSLFN and EE blocks significantly lowers performance with modest reduction in size, highlighting BEExformer’s efficiency. Among binarized Transformers like BinaryBERT, BiBERT and BiT; BEExformer offers a superior performance-size trade-off. While quantized models like TernaryBERT and Q-BERT achieve competitive performance, they come at the expense of much larger model sizes.

In Figure \ref{fig3b}, BEExformer (Proposed) 1-1, positioned on the optimal Pareto front, achieves a 21.30 $\times$ model size reduction compared to its full-precision counterpart (BEExformer FP) while maintaining competitive performance (3.51\% lower). This underscores the effectiveness of BAT in model efficiency. Other ablations also fall on the second front but offer much less performance given their model sizes, highlighting the contributions of SLFN and EE. Among SOTA models, PABEE 32-32, also on the optimal front, outperforms BEExformer 1-1 by 2.65\%, but at 3.22 $\times$ the size, making it less efficient. Larger models like BE3R\_Electra 32-32, LayerDrop Large 32-32, and ElasticBERT Large 32-32 achieve higher performance but with substantial size increases, while their base variants witness a performance drop due to fewer layers.

\begin{figure*}[h!]
    \centering
    \begin{subfigure}[h]{0.3\textwidth}
        \centering
        \includegraphics[width=\linewidth, height=3.6cm]{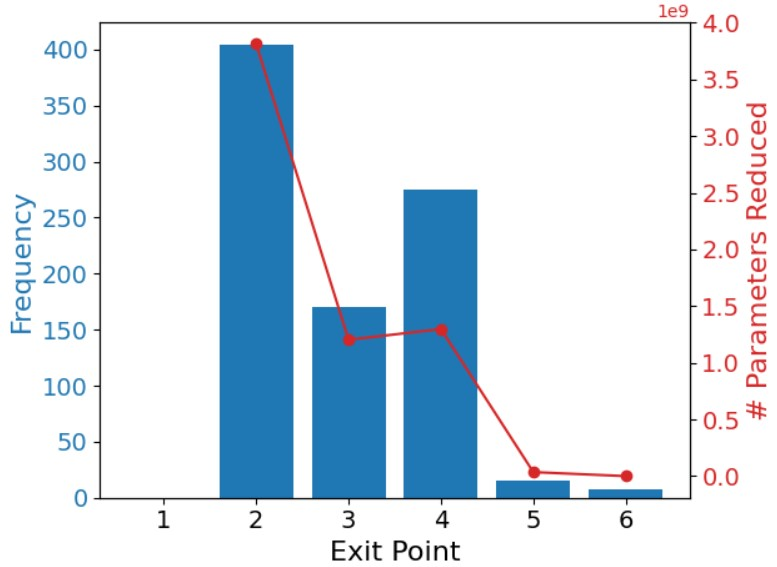}
        \caption{Dataset: SST-2}
        \label{fig2a}       
    \end{subfigure}%
    ~
    \begin{subfigure}[h]{0.3\textwidth}
        \centering
        \includegraphics[width=\linewidth, height=3.6cm]{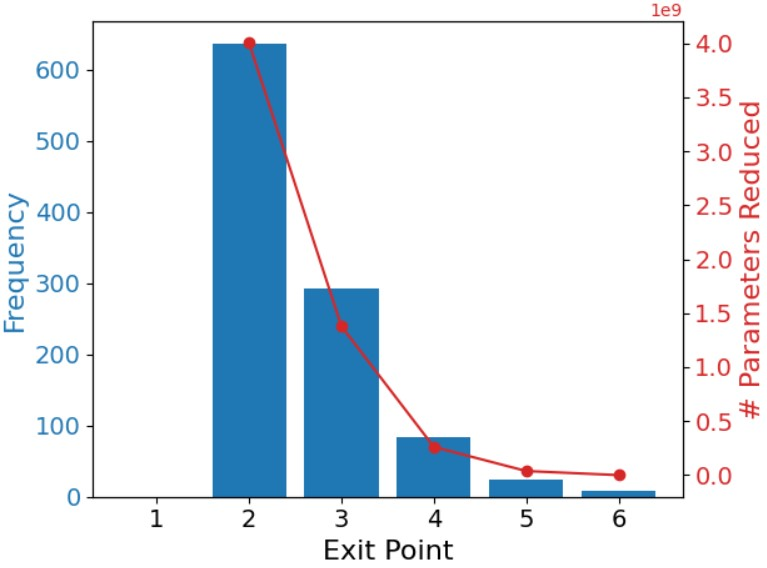}
        \caption{Dataset: CoLA}
        \label{fig2b}       
    \end{subfigure}
    ~
    \begin{subfigure}[h]{0.3\textwidth}
        \centering
        \includegraphics[width=\linewidth, height=3.6cm]{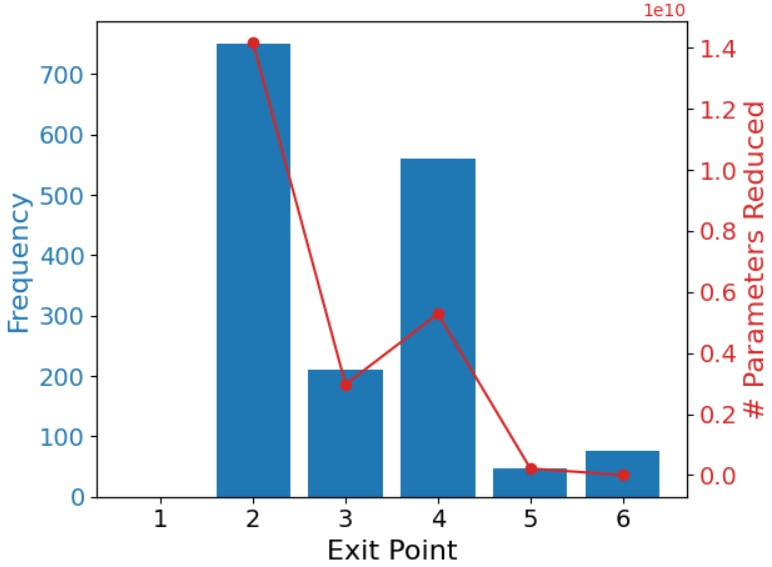}
        \caption{Dataset: MRPC}
        \label{fig2c}       
    \end{subfigure}%
     
    \begin{subfigure}[h]{0.3\textwidth}
        \centering
        \includegraphics[width=\linewidth, height=3.6cm]{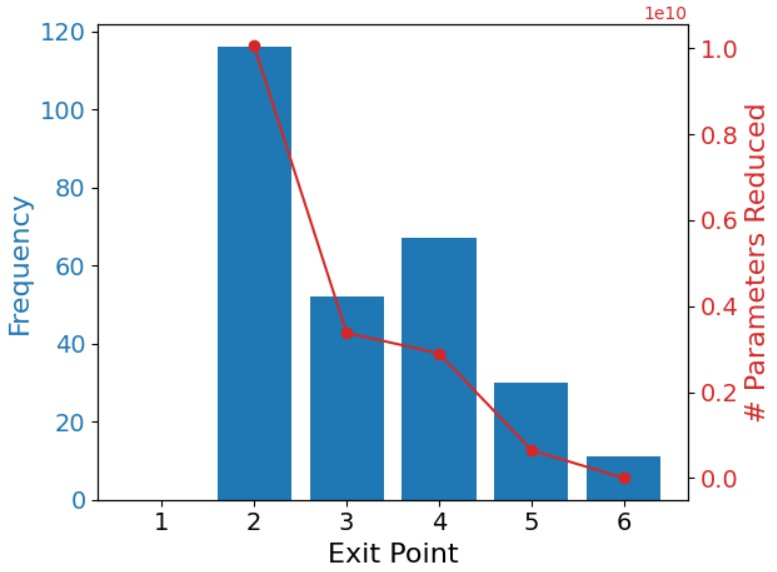}
        \caption{Dataset: RTE}
        \label{fig2d}       
    \end{subfigure}
    ~
    \begin{subfigure}[h]{0.3\textwidth}
    \centering
    \includegraphics[width=\linewidth, height=3.6cm]{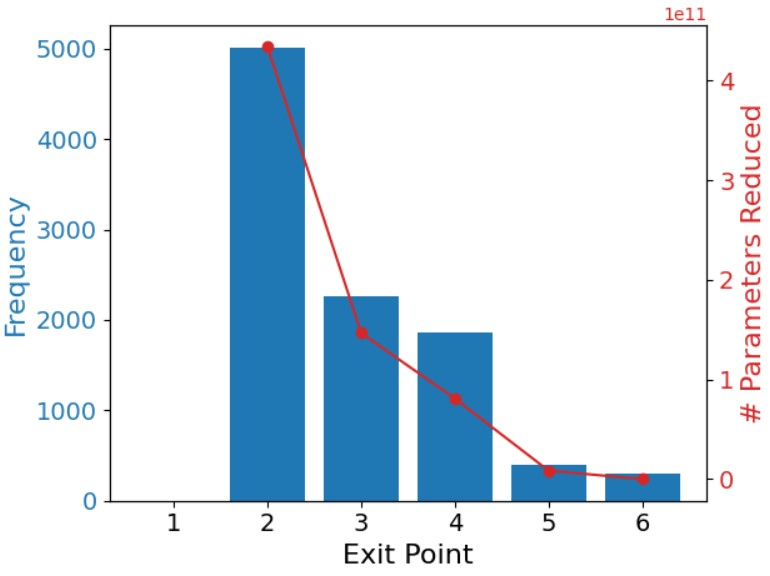}
    \caption{Dataset: MNLI-m}
    \label{fig2e}       
    \end{subfigure}%
    ~ 
    \begin{subfigure}[h]{0.3\textwidth}
        \centering
        \includegraphics[width=\linewidth, height=3.6cm]{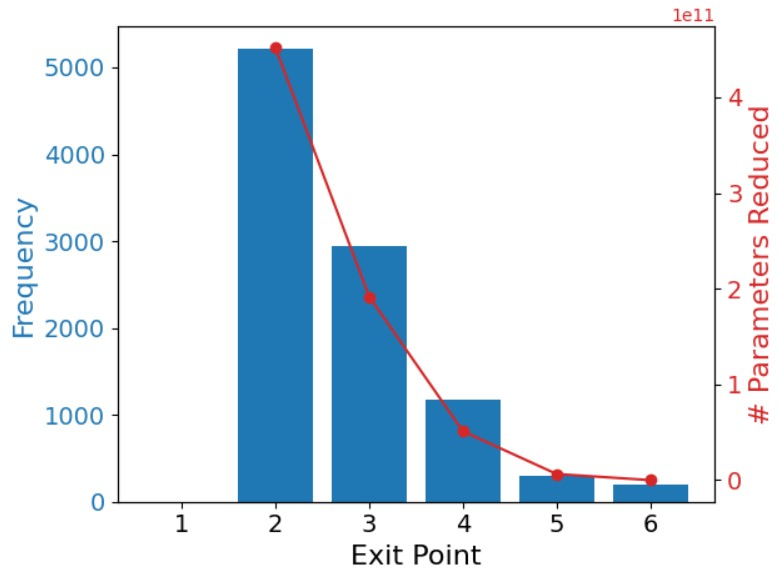}
        \caption{Dataset: MNLI-mm}
        \label{fig2f}       
    \end{subfigure}

    \begin{subfigure}[h]{0.3\textwidth}
        \centering
        \includegraphics[width=\linewidth, height=3.6cm]{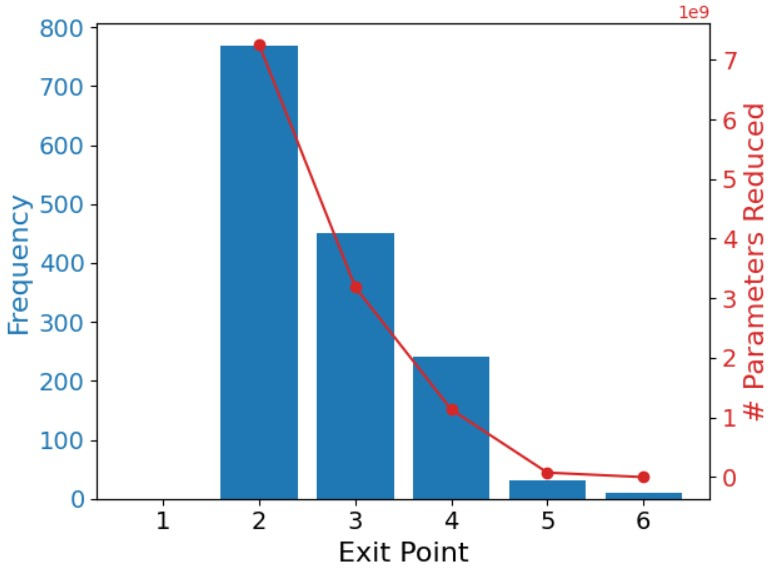}
        \caption{Dataset: STS-B}
        \label{fig2g}       
    \end{subfigure}
    ~
    \begin{subfigure}[h]{0.3\textwidth}
    \centering
    \includegraphics[width=\linewidth, height=3.6cm]{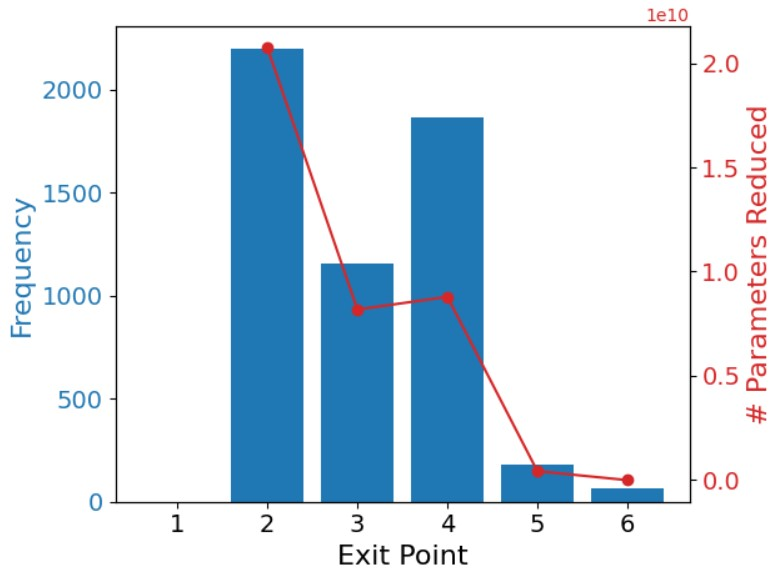}
    \caption{Dataset: QNLI}
    \label{fig2h}       
    \end{subfigure}%
    ~ 
    \begin{subfigure}[h]{0.3\textwidth}
        \centering
        \includegraphics[width=\linewidth, height=3.6cm]{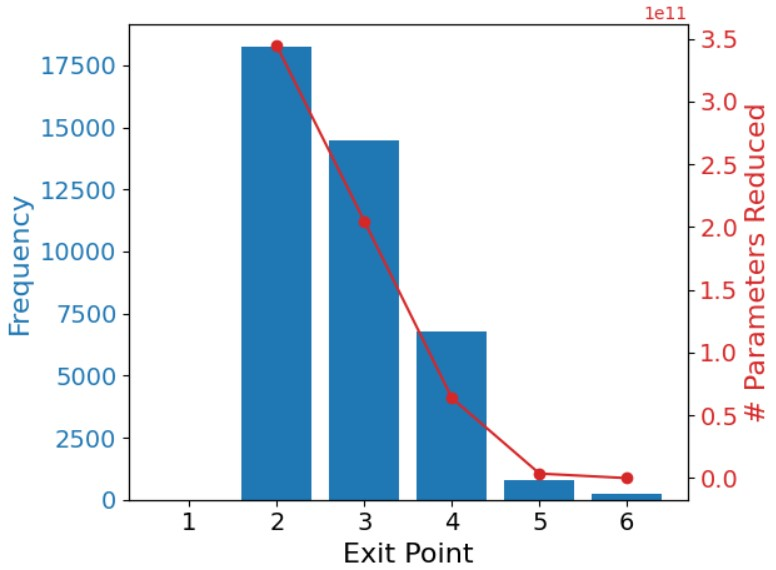}
        \caption{Dataset: QQP}
        \label{fig2i}       
    \end{subfigure}
    \caption{Distribution of exits and total number of parameters saved from computation during inference.}
    \label{fig2}
    
\end{figure*}

\begin{table}[t!]
  \begin{center}
    \caption{Percentage Reduction in FLOPs due to EE}
    \label{table:table3}
    \setlength{\tabcolsep}{0.6em} 
    \renewcommand{\arraystretch}{1} 
 \begin{tabular}{|l|l|l|l|} 
 \hline
 Dataset & WEE (GFLOPs) $\downarrow$  & EE (GFLOPs) $\downarrow$ &  Reduction (\%) $\uparrow$
\\ 
 \hline
 SST-2 & 0.240 & 0.116 & 51.43\% \\
 MRPC & 0.240 & 0.123 & 48.71\% \\
 CoLA & 0.240 & 0.101 & 57.69\% \\
 RTE & 0.242 & 0.128 & 47.34\% \\
 MNLI-m & 0.242 & 0.115 & 52.50\% \\
 MNLI-mm & 0.242 & 0.109 & 54.84\% \\
 STS-B & 0.240 & 0.108 & 54.82\% \\
 QNLI & 0.240 & 0.122 & 49.30\% \\
 QQP & 0.240 & 0.111 & 53.83\% \\
 \hline
 Average & 0.241 & 0.115 & 52.27\% \\
 \hline
 \multicolumn{4}{p{0.63\linewidth}}{Note- EE: Applying the  proposed EE; WEE: Without EE\newline
 GFLOPs is rounded to three decimal places, whereas the percentage reduction is calculated using all significant digits for accurate measurement. \newline
 $\uparrow$ indicates higher value is desired while $\downarrow$ indicates lower value is desired.
 }
\end{tabular}
\end{center}
\end{table}

\subsection{Distribution of Exit Points}
\label{de}
For a deeper insight, we plot the exit point distribution along with the number of parameters saved, i.e. not computed due to EE, during inference over all datasets in Figure \ref{fig2}. Overall, 48.39\% of the observations exit after the second transformer block, while a mere 1.96\% samples need to go through the last transformer block. This highlights savings in computation during inference compared to traditional models, which involve all layers of the network. The total number of parameters saved from computation during inference is proportional to how early the exit condition is satisfied. Its maximum effect is noticed at the second exit and then starts decreasing. For subsequent exits, the plots indicate that despite having a higher frequency in some cases, their impact on saving parameters might not be that pronounced. An interesting observation is that none of the samples exits from the first block during inference. This is due to the substantial reduction in entropy from the initialized value $\ln(m)$ after execution through the first transformer block.

Furthermore, Table \ref{table:table3} presents the percentage reduction in overall FLOPs during inference over all datasets compared to the variant without EE. On average, the EE mechanism in BEExformer saves around 52.27\% FLOPs during inference accompanied with 3.22\% increase in performance. It affirms our motivation behind EE to utilize fewer transformer blocks than in the full architecture during inference.

\subsection{Effect of EE Threshold}
\label{ed}

\begin{figure*}[h!]
    \centering
    \begin{subfigure}[h]{0.5\linewidth}
        \centering
        \includegraphics[width=\linewidth]{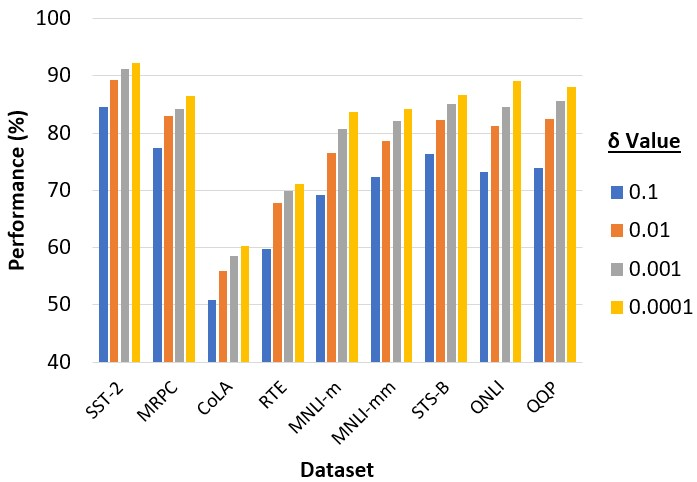}
        \caption{Comparison of performance}
        \label{fig6a}       
    \end{subfigure}%
    ~
    \begin{subfigure}[h]{0.5\linewidth}
        \centering
        \includegraphics[width=\linewidth]{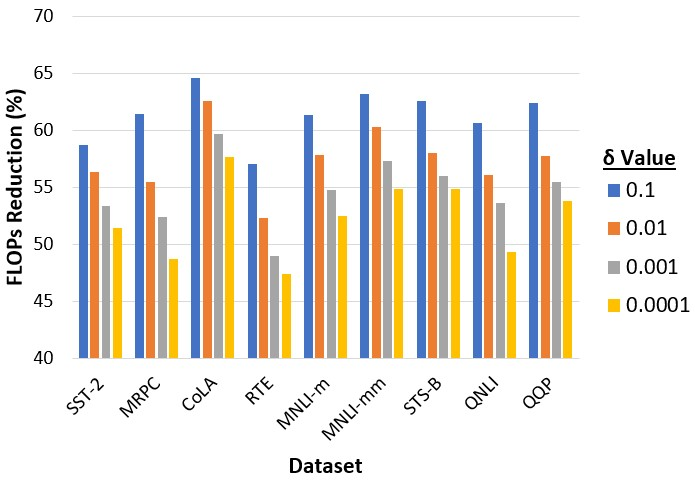}
        \caption{Comparison of percentage reduction in FLOPs}
        \label{fig6b}       
    \end{subfigure}

    \caption{Comparison of the quantized models for varying $\delta$ on all the datasets. Here, $\delta=0.0001$ corresponds to the proposed version of BEExformer.}
    
    \label{fig6}
    
\end{figure*}

We observe that the EE threshold ($\delta$) plays a pivotal role in determining the trade-off between the efficacy of the results and the reduction in FLOPs during inference. Figure \ref{fig6} demonstrates the rationale for selecting $\delta = 0.0001$ in the proposed BEExformer through performance comparison for $\delta \in \{0.1, 0.01, 0.001, 0.0001\}$. Relative to $\delta=0.0001$, the overall GFLOPs during inference decrease by 9.07\%, 5.14\% and 2.34\% for $\delta$ values of 0.1, 0.01 and 0.001 respectively. However, it translates into a performance drop by 11.61\%, 4.98\% and 2.22\% for the same set of $\delta$ values. Therefore, larger $\delta$ values leads to faster inference due to reduced GFLOPs, accompanied with a significant drop in performance as confidence in the output decreases. This effect is more prominent for $\delta=0.1$ and fades with lower $\delta$ values. Alternatively, for smaller values of $\delta$, the confidence in outputs increases with a slight delay in inference. Thus, setting $\delta \to 0^{+}$ proves to be a win-win situation leading to a higher performance along with reduced FLOPs for inference. It tackles ``overthinking" wherein the entropy of logits starts rising instead of decreasing. Therefore, letting the inputs with even a slight reduction in entropy be processed further, otherwise execution is terminated.

\begin{figure}[h!]
    \centering
    \begin{subfigure}[h]{0.5\linewidth}
        \centering
        \includegraphics[width=0.72\linewidth]{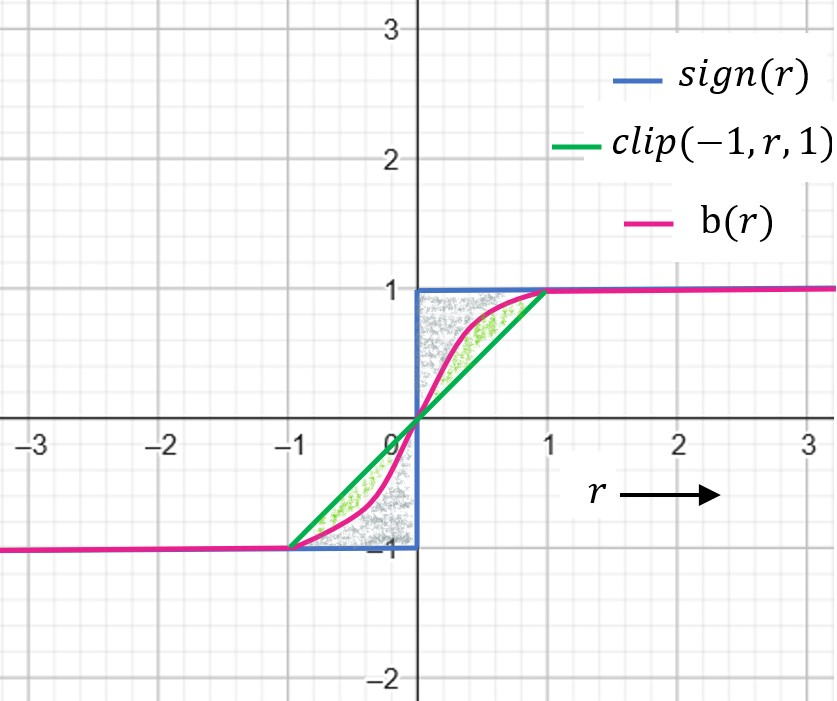}
        \caption{Comparison $b(r)$ with equivalent functions}
        \label{fig4a}       
    \end{subfigure}%
    ~
    \begin{subfigure}[h]{0.5\linewidth}
        \centering
        \includegraphics[width=0.72\linewidth]{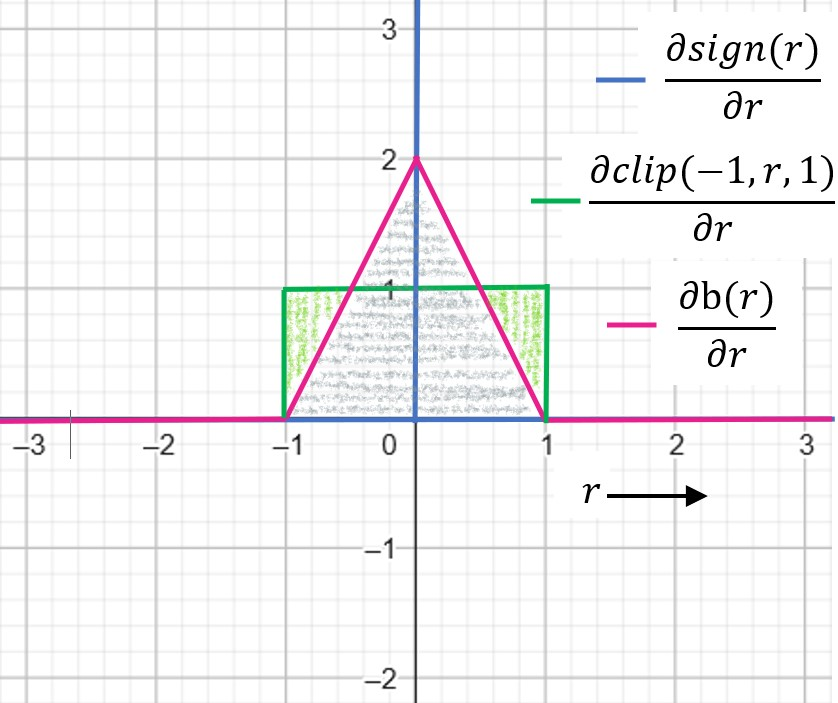}
        \caption{Comparison of derivatives of $b(r)$ with equivalent functions}
        \label{fig4b}       
    \end{subfigure}

    \caption{Comparison of plots for $sign(r)$, $clip(-1,r,1)$, and the proposed binarization function $b(r)$ along with their derivatives. The shaded areas indicate the difference between the functions.}
    
    \label{fig4}
    
\end{figure}

\begin{figure}[h!]
    \centering
    \includegraphics[width=0.6\linewidth, height=5.7cm]{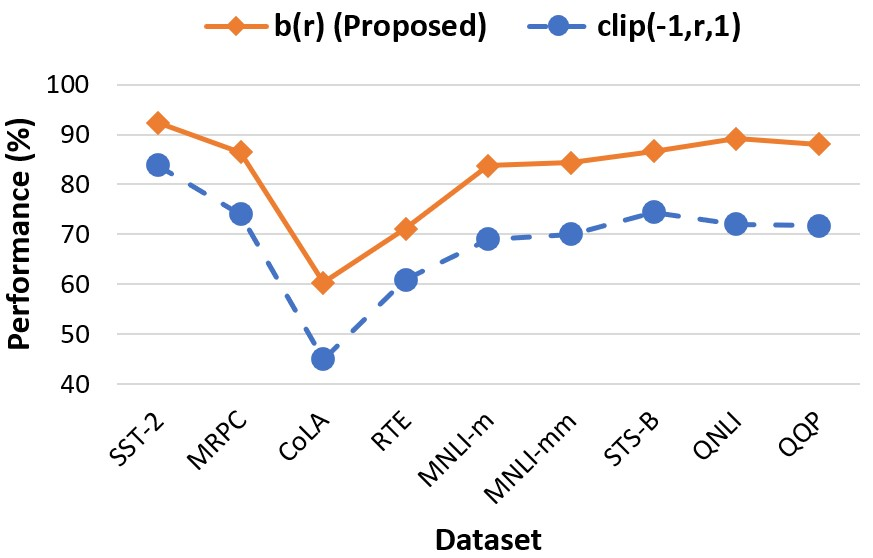}
    \caption{Comparison of performance of the proposed binarization function $b(r)$ with $clip(-1,r,1)$ function on all the datasets.}

    \label{fig7}
    
\end{figure}

\subsection{Comparison of Binarization Functions}
\label{bd}
To assess the effectiveness of the binarization function $b(r)$ in BEExformer, it has been compared with $clip(-1,r,1)$ \cite{Ref18}. To give a clear picture, Figure \ref{fig4} graphically plots $sign(r)$, $clip(-1,r,1)$ and $b(r)$ along with their derivatives with shaded areas highlighting the difference among them. While $b(r)$ improves gradient stability by providing a smoother approximation to $sign(r)$, $clip(-1,r,1)$ lacks the smooth transition provided by $b(r)$, potentially leading to sharper gradient updates and instability in certain cases. Furthermore, $b(r)$ is a tighter approximation to $sign(r)$ compared to $clip(-1,r,1)$, leading to a more accurate gradient estimation. This gets reflected in the performance too when $b(r)$ is substituted with $clip(-1,r,1)$ in the proposed BEExformer, where $b(r)$ surpasses $clip(-1,r,1)$ by 13.48\% across all datasets as illustrated in Figure \ref{fig7}.

\newpage

\section{Conclusion}
\label{cl}
In this paper, BEExformer-- a binarized transformer architecture with SLFN and EE for textual content, has been proposed. It incorporates a differentiable piecewise approximation function for BAT to ensure that the gradient computation accounts for both the magnitude and the sign of real-valued weights. A binarized SLFN is integrated in each transformer block, for selective retention of significant information. Overall, it accomplishes 21.30 $\times$ reduction in size with minimal impact on performance compared to a full-precision model. Furthermore, it accelerates inference through an intuitive EE technique monitoring entropy changes in logits. By applying soft-routing loss during training, the losses from all exits are accounted to optimally update the weights of each transformer block. This not only reduces the overall FLOPs during inference by 52.27\%, but also offers a solution to the ``overthinking" problem. Extensive evaluation on nine benchmark datasets showcases its ability to deliver Pareto-optimal results concerning both efficacy as well as efficiency. In addition, the efficacy of each component in BEExformer has been validated through various ablation studies. This makes BEExformer an apt choice for applications with a limited computational budget combined with the demand for faster inference. However, the proposed architecture has certain limitations as well. Being a dynamic architecture, the inference time and FLOPs cannot be predicted exactly for individual samples, but are bounded between that required for executing a minimum of two and a maximum of six transformer blocks (as observed in Section \ref{de}). The current version is based on the transformer encoder and is limited to Natural Language Understanding (NLU) tasks only. In the future, the encoder blocks in the BEExformer could be augmented with decoder blocks providing autoregressive capabilities to perform Natural Language Generation (NLG) tasks. This would require redesigning it into a binarized encoder‑decoder architecture, with multiple pathways between the encoder and decoder blocks, each equipped with auxiliary blocks for EE. This would yield a lightweight model with reduced precision and faster inference as in BEExformer, while extending its capabilities to NLG tasks.

\section*{Acknowledgments}
This study is supported by the University Grants Commission (UGC), India, through a research fellowship awarded to Wazib Ansar for his Ph.D. pursuit. However, the UGC played no role in the study’s design, execution, analysis or decision to publish the findings.


\bibliographystyle{unsrt}  

\begin{thebibliography}{34}


\bibitem{li2022survey}
Q. Li et al., “A survey on text classification: From traditional to deep learning”, ACM Transactions on Intelligent Systems and Technology (TIST), vol. 13, no. 2, pp. 1–41, 2022.

\bibitem{Ref41}
L. Gu, W. Zhang, Z. Wang, D. Zeng, and H. Jin, "Service management and energy scheduling toward low-carbon edge computing," IEEE Transactions on Sustainable Computing, vol. 8, no. 1, pp. 109–119, 2022.

\bibitem{Ref42}
S. M. Hasan, T. Islam, M. Saifuzzaman, K. R. Ahmed, C.-H. Huang, and A. R. Shahid, "Carbon emission quantification of machine learning: A review," IEEE Transactions on Sustainable Computing, 2025.

\bibitem{Ref43}
M. Alswaitti, R. Verdecchia, G. Danoy, P. Bouvry, and J. Pecero, "Training green AI models using elite samples," IEEE Transactions on Sustainable Computing, 2025.

\bibitem{wa2025survey}
W. Ansar, S. Goswami, and A. Chakrabarti, "From Transformers to LLMs: A Systematic Survey of Efficiency Considerations in NLP," arXiv:2406.16893, 2025. [Online]. Available: https://arxiv.org/abs/2406.16893

\bibitem{Ref21}
F. Lagunas, E. Charlaix, V. Sanh, and A. M. Rush, “Block Pruning For Faster Transformers”, in Proceedings of the 2021 Conference on Empirical Methods in Natural Language Processing, 2021, pp. 10619–10629.

\bibitem{Ref22}
V. Sanh, “DistilBERT, a distilled version of BERT: smaller, faster, cheaper and lighter”, arXiv preprint arXiv:1910. 01108, 2019.

\bibitem{Ref23}
S. Wang, B. Z. Li, M. Khabsa, H. Fang, and H. Ma, “Linformer: Self-attention with linear complexity”, arXiv preprint arXiv:2006. 04768, 2020.

\bibitem{Ref6}
S. Mangrulkar, A. Ms, and V. Sembium, “Be3r: Bert based early-exit using expert routing”, in Proceedings of the 28th ACM SIGKDD Conference on Knowledge Discovery and Data Mining, 2022, pp. 3504–3512.

\bibitem{Ref26}
Y. Kaya, S. Hong, and T. Dumitras, “Shallow-deep networks: Understanding and mitigating network overthinking”, in International conference on machine learning, 2019, pp. 3301–3310.

\bibitem{Ref27}
J. Xin, R. Nogueira, Y. Yu, and J. Lin, “Early exiting BERT for efficient document ranking”, in Proceedings of SustaiNLP: Workshop on Simple and Efficient Natural Language Processing, 2020, pp. 83–88.

\bibitem{Ref12}
S. Shen et al., “Q-bert: Hessian based ultra low precision quantization of bert”, in Proceedings of the AAAI Conference on Artificial Intelligence, 2020, vol. 34, pp. 8815–8821.

\bibitem{Ref29}
Z. Yao, R. Yazdani Aminabadi, M. Zhang, X. Wu, C. Li, and Y. He, “Zeroquant: Efficient and affordable post-training quantization for large-scale transformers”, Advances in Neural Information Processing Systems, vol. 35, pp. 27168–27183, 2022.

\bibitem{Ref19}
Z. Liu, W. Luo, B. Wu, X. Yang, W. Liu, and K.-T. Cheng, “Bi-real net: Binarizing deep network towards real-network performance”, International Journal of Computer Vision, vol. 128, pp. 202–219, 2020.

\bibitem{Ref24}
M. W. U. Rahman et al., “Quantized transformer language model implementations on edge devices”, in 2023 International Conference on Machine Learning and Applications (ICMLA), 2023, pp. 709–716.

\bibitem{rokh2023comprehensive}
B. Rokh, A. Azarpeyvand, and A. Khanteymoori, “A comprehensive survey on model quantization for deep neural networks in image classification”, ACM Transactions on Intelligent Systems and Technology, vol. 14, no. 6, pp. 1–50, 2023.

\bibitem{Ref30}
H. Shen, N. Mellempudi, X. He, Q. Gao, C. Wang, and M. Wang, “Efficient post-training quantization with fp8 formats”, Proceedings of Machine Learning and Systems, vol. 6, pp. 483–498, 2024.

\bibitem{Ref31}
G. Shomron, F. Gabbay, S. Kurzum, and U. Weiser, “Post-training sparsity-aware quantization”, Advances in Neural Information Processing Systems, vol. 34, pp. 17737–17748, 2021.

\bibitem{Ref25}
M. Chen et al., “Efficientqat: Efficient quantization-aware training for large language models”, arXiv preprint arXiv:2407. 11062, 2024.

\bibitem{qin2023bibench}
H. Qin et al., “Bibench: Benchmarking and analyzing network binarization”, in International Conference on Machine Learning, 2023, pp. 28351–28388.

\bibitem{Ref18}
I. Hubara, M. Courbariaux, D. Soudry, R. El-Yaniv, and Y. Bengio, “Binarized neural networks”, Advances in neural information processing systems, vol. 29, 2016.

\bibitem{Ref1}
H. Bai et al., “BinaryBERT: Pushing the Limit of BERT Quantization”, in Proceedings of the 59th Annual Meeting of the Association for Computational Linguistics and the 11th International Joint Conference on Natural Language Processing (Volume 1: Long Papers), 2021, pp. 4334–4348.

\bibitem{Ref2}
H. Qin et al., “BiBERT: Accurate Fully Binarized BERT”, in International Conference on Learning Representations.

\bibitem{Ref3}
Z. Liu et al., “Bit: Robustly binarized multi-distilled transformer”, Advances in neural information processing systems, vol. 35, pp. 14303–14316, 2022.

\bibitem{Ref28}
S. Stanton, P. Izmailov, P. Kirichenko, A. A. Alemi, and A. G. Wilson, “Does knowledge distillation really work?”, Advances in Neural Information Processing Systems, vol. 34, pp. 6906–6919, 2021.

\bibitem{Ref46}
K. Zeng, Z. Wan, H. Gu, and T. Shen, “Self-knowledge distillation enhanced binary neural networks derived from underutilized information,” Applied Intelligence, vol. 54, no. 6, pp. 4994–5014, Mar. 2024, doi: 10.1007/s10489-024-05444-8.

\bibitem{Ref44}
Y. Han, G. Huang, S. Song, L. Yang, H. Wang, and Y. Wang, "Dynamic neural networks: A survey," IEEE Transactions on Pattern Analysis and Machine Intelligence, vol. 44, no. 11, pp. 7436–7456, 2021.

\bibitem{Ref40}
B. Fu, F. Chen, P. Li, and D. Zeng, "Serving Transformer Models via Joint Request Scheduling and Batching in the Network Edge," IEEE Transactions on Sustainable Computing, 2025.

\bibitem{Ref4}
W. Zhou, C. Xu, T. Ge, J. McAuley, K. Xu, and F. Wei, “Bert loses patience: Fast and robust inference with early exit”, Advances in Neural Information Processing Systems, vol. 33, pp. 18330–18341, 2020.

\bibitem{Ref5}
J. Xin, R. Tang, J. Lee, Y. Yu, and J. Lin, “DeeBERT: Dynamic Early Exiting for Accelerating BERT Inference”, in Proceedings of the 58th Annual Meeting of the Association for Computational Linguistics, 2020, pp. 2246–2251.

\bibitem{Ref14}
X. Liu et al., “Towards Efficient NLP: A Standard Evaluation and A Strong Baseline”, in Proceedings of the 2022 Conference of the North American Chapter of the Association for Computational Linguistics: Human Language Technologies, 2022, pp. 3288–3303.

\bibitem{Ref15}
W. Ansar, S. Goswami, A. Chakrabarti, and B. Chakraborty, “A novel selective learning based transformer encoder architecture with enhanced word representation”, Applied Intelligence, vol. 53, no. 8, pp. 9424–9443, 2023.

\bibitem{Ref13}
W. Zhang et al., “TernaryBERT: Distillation-aware Ultra-low Bit BERT”, in Proceedings of the 2020 Conference on Empirical Methods in Natural Language Processing (EMNLP), 2020, pp. 509–521.

\bibitem{fan2019reducing}
A. Fan, E. Grave, and A. Joulin, “Reducing transformer depth on demand with structured dropout”, arXiv preprint arXiv:1909. 11556, 2019.

\bibitem{michel2019sixteen}
P. Michel, O. Levy, and G. Neubig, “Are sixteen heads really better than one?”, Advances in neural information processing systems, vol. 32, 2019.

\bibitem{Ref17}
A. Vaswani, “Attention is all you need”, Advances in Neural Information Processing Systems, 2017.

\bibitem{Ref7}
A. Wang, “Glue: A multi-task benchmark and analysis platform for natural language understanding”, arXiv preprint arXiv:1804. 07461, 2018.

\bibitem{devlin2019bert}
J. Devlin, M.-W. Chang, K. Lee, and K. Toutanova, “Bert: Pre-training of deep bidirectional transformers for language understanding”, in Proceedings of the 2019 conference of the North American chapter of the association for computational linguistics: human language technologies, volume 1 (long and short papers), 2019, pp. 4171–4186.

\bibitem{lan2019albert}
Z. Lan, M. Chen, S. Goodman, K. Gimpel, P. Sharma, and R. Soricut, “Albert: A lite bert for self-supervised learning of language representations”, arXiv preprint arXiv:1909. 11942, 2019.

\bibitem{liu2019roberta}
Y. Liu et al., “Roberta: A robustly optimized bert pretraining approach”, arXiv preprint arXiv:1907. 11692, 2019.

\bibitem{Ref45}
J. Kaplan, S. McCandlish, T. Henighan, T.B. Brown, B. Chess, R. Child, S. Gray, A. Radford, J. Wu, and D. Amodei, “Scaling Laws for Neural Language Models”, arXiv preprint arXiv:2001.08361 (2020).

\end{thebibliography}

\end{document}